%% file: main.tex
\documentclass[sigconf,nonacm]{acmart}
\setcopyright{none}
\usepackage{booktabs}
\usepackage{tabularx}
\usepackage{xspace}
\usepackage{algorithm}
\usepackage{algpseudocode}

\begin{document}

\title{From Test-Time Scaling to Reusable Memory: Measuring Crystallization in Text-to-SQL}

\author{Jiaqian Wang}
\affiliation{%
  \institution{Xidian University}
  \city{Xi'an, Shaanxi}
  \country{China}}

\author{Yutao Qi}
\authornote{Corresponding author.}
\affiliation{%
  \institution{Xidian University}
  \city{Xi'an, Shaanxi}
  \country{China}}

\author{Wenjin Hou}
\affiliation{%
  \institution{Xidian University}
  \city{Xi'an, Shaanxi}
  \country{China}}

\author{Yuanxi Che}
\affiliation{%
  \institution{Xidian University}
  \city{Xi'an, Shaanxi}
  \country{China}}

\author{Muning Wen}
\affiliation{%
  \institution{Shanghai Jiao Tong University}
  \department{School of Artificial Intelligence}
  \city{Shanghai}
  \country{China}}

\begin{abstract}
Test-time scaling can correct difficult text-to-SQL queries, but the extra
computation is normally discarded after each answer. Systems increasingly
retain verified repair episodes, yet evaluations still report one end-to-end
score. It cannot distinguish replay on recurring questions from help on unseen
questions, or identify the responsible memory choice. We call measuring this
future value the \emph{crystallization problem}. Our controlled evaluation
holds the single-shot solver fixed and varies one memory choice at a time. We
separately measure replay, cross-question retention, and held-out same-database
transfer. On BIRD, storing verified corrected queries improves held-out
first-attempt accuracy by $4.34$ percentage points. This gain captures $44.4\%$
of the accuracy headroom provided by on-demand repair on the same questions.
Controlled interventions identify database-specific content as the main
operating ingredient. Reliable verification and broader retrieval coverage yield
supported gains; richer formats and elaborate retrievers do not. Open-source
code, evaluation artifacts, and reproduction instructions are available at
\href{https://github.com/ai-jiaqian/text-to-sql-memory-crystallization}{github.com/ai-jiaqian/text-to-sql-memory-crystallization}.
\end{abstract}


\maketitle

\input{sections/intro}

\input{sections/related}
\input{sections/framework}
\input{sections/setup}
\input{sections/results}
\input{sections/analysis}
\input{sections/threats}
\input{sections/conclusion}

\bibliographystyle{ACM-Reference-Format}
\bibliography{references}

\appendix
\input{sections/x_extended}

\end{document}

%% file: sections/intro.tex
\section{Introduction}
\label{sec:intro}

\input{figures/fig1_framework}

Text-to-SQL lets users query a relational database in natural
language~\cite{spider2018,bird2023}, and large language models have pushed
first-attempt accuracy to a high
level~\cite{dinsql2023,dailsql2024,chasesql2025,xiyansql2025}. A failed
first attempt is not the end of the road. Additional inference-time
computation can sample more candidates, aggregate them by
self-consistency~\cite{wang2023selfconsist}, or repair the query against
execution
feedback~\cite{macsql2025,chen2024selfdebug,snell2024tts,zhang2025ttssurvey}. On BIRD, an
execution-guided repair loop with a correctness signal lifts a strong
open-weight model from about $62\%$ to about $72\%$ within three repair
rounds. The catch is that this improvement is purchased separately for every
failed question. The computation benefits only the question it is spent on, and the
experience accumulated during repair is discarded once the answer is
delivered.

Since that experience was bought with real computation, a natural idea is to
keep it. The information produced while correcting one question is a test-time
episode. Once verified, it becomes a card stored in a bank for that database.
A later question can then retrieve relevant cards instead of starting from
scratch (Figure~\ref{fig:teaser}).

Whether this memory pays off depends on what arrives next. If the same repaired
question returns, its card can supply the stored solution---\emph{exact-query
replay}. If a new question arrives, the bank helps only when experience crosses
question boundaries; we measure this on held-out questions as \emph{transfer}.
Between the two, we use a stricter diagnostic: remove a repaired question's own
card and ask whether the remaining cards can still help re-answer it. We call
this \emph{cross-question retention}. Replay and transfer correspond to the
recurring and new questions in a workload; retention isolates the ability to
reuse experience across questions. The recurring/new mix determines whether
the computation spent building memory pays back.

Turning test-time computation into memory is already a practiced route. For
general LLM agents the idea is explicit: reasoning strategies, solutions and
insights, workflows, and experiential lessons are all distilled from test-time
trajectories into reusable
memory~\cite{reasoningbank2025,suzgun2026cheatsheet,wang2025awm,zhao2024expel}.
Self-evolving text-to-SQL
systems can be read as the same route instantiated on this task. They retain
corrected queries and worked examples~\cite{memosql2026,lpesql2024},
self-correction guidelines and optimized hints~\cite{magic2025,tahoe2026},
semantic agent memories~\cite{agentsm2026,merit2026}, and distilled domain
knowledge~\cite{orange2025,katsql2025}, and they consistently report
end-to-end gains. That memory can help is no longer in question.

What still needs asking is \emph{how} it helps, a question on which
existing evaluations are largely silent. In this line of work the loop is almost always scored
as a whole, with a single end-to-end number. That number conflates two
different things. One is where the gain goes: a recurring question answered
cheaply again is indistinguishable from memory helping a question
never seen before. The other is where the gain comes from: how episodes are
collected, verified, stored, and retrieved often changes together with the
surrounding pipeline. No single choice can then be credited. The agent-memory community
itself warns that one end-to-end score hides how a memory actually
behaves~\cite{memoryagentbench2025,gao2026survey,shao2026misevolve}, and
attribution there often rests on comparatively weak signals such as LLM-based
judgments. In practice, an operator cannot tell whether collecting episodes
will pay off for a workload, nor a researcher which component makes it pay
off.

This is the \emph{crystallization problem}: how do verified test-time episodes
create value for future questions over the same database? Our controlled
evaluation separates where that value appears from which memory choice
produces it under a fixed solver, following studies that isolate which
demonstration components carry observed gains~\cite{min2022demonstrations}.
We ask three questions.
\textbf{RQ1} measures how much value appears through replay,
cross-question retention, and held-out transfer, and compares transfer with
the repair headroom on the same questions. \textbf{RQ2} asks what information
carries transfer: generic examples, aligned question--SQL pairs, or exposure to
the target database. \textbf{RQ3} tests which choices in collecting, verifying,
formatting, and retrieving episodes materially change the results.

Execution-guided repair supplies the episodes in our main instantiation. We do
not assume that repaired episodes transfer better than other episodes; RQ3
tests that possibility.
Replay and retention are evaluated on collection questions, while transfer is
evaluated on held-out questions from the same databases. CR is the held-out
transfer lift divided by the on-demand repair headroom on those same questions.
It is a descriptive ratio, not a probability. The card writer never sees gold
SQL (Section~\ref{sec:framework-writer}). Main comparisons use three paired
seeds and account for variation across databases and questions.
Figure~\ref{fig:protocol} summarizes the protocol.

On BIRD, a verified, database-scoped bank that stores corrected episodes
verbatim delivers value in all three settings. Among verified-repaired
collection questions, it replays nearly all stored answers and transfers to new questions over the same
databases: held-out first-attempt accuracy rises by $4.34$ percentage points,
which is $44.4\%$ of the on-demand repair headroom measured on the same
questions. Mechanism interventions show that transfer is carried mainly by
having seen the target database. Local cards expose useful identifiers, values,
joins, and query structures even when their question--SQL pairings are
shuffled. Reliable verification and broad retrieval coverage produce clear
gains. Richer card formats, more elaborate retrievers, and repair-specific
content do not show a statistically supported advantage.

Our contributions are:
\begin{itemize}
  \item \textbf{Formalizing the crystallization problem} as an operational
  measurement problem. CR compares held-out memory lift with on-demand repair
  headroom under the same solver, data split, and repair budget.
  \item \textbf{Separated measurement of future value:} exact-query replay,
  cross-question retention, and held-out same-database transfer are evaluated
  separately, so a reported gain maps to a concrete future-use setting.
  \item \textbf{Controlled attribution:} with a fixed solver and
  paired comparisons that change one choice at a time, we test three
  explanations of transfer directly rather than comparing bundled systems.
\end{itemize}

%% file: figures/fig1_framework.tex
\begin{figure}[t]
\centering
\includegraphics[width=\columnwidth]{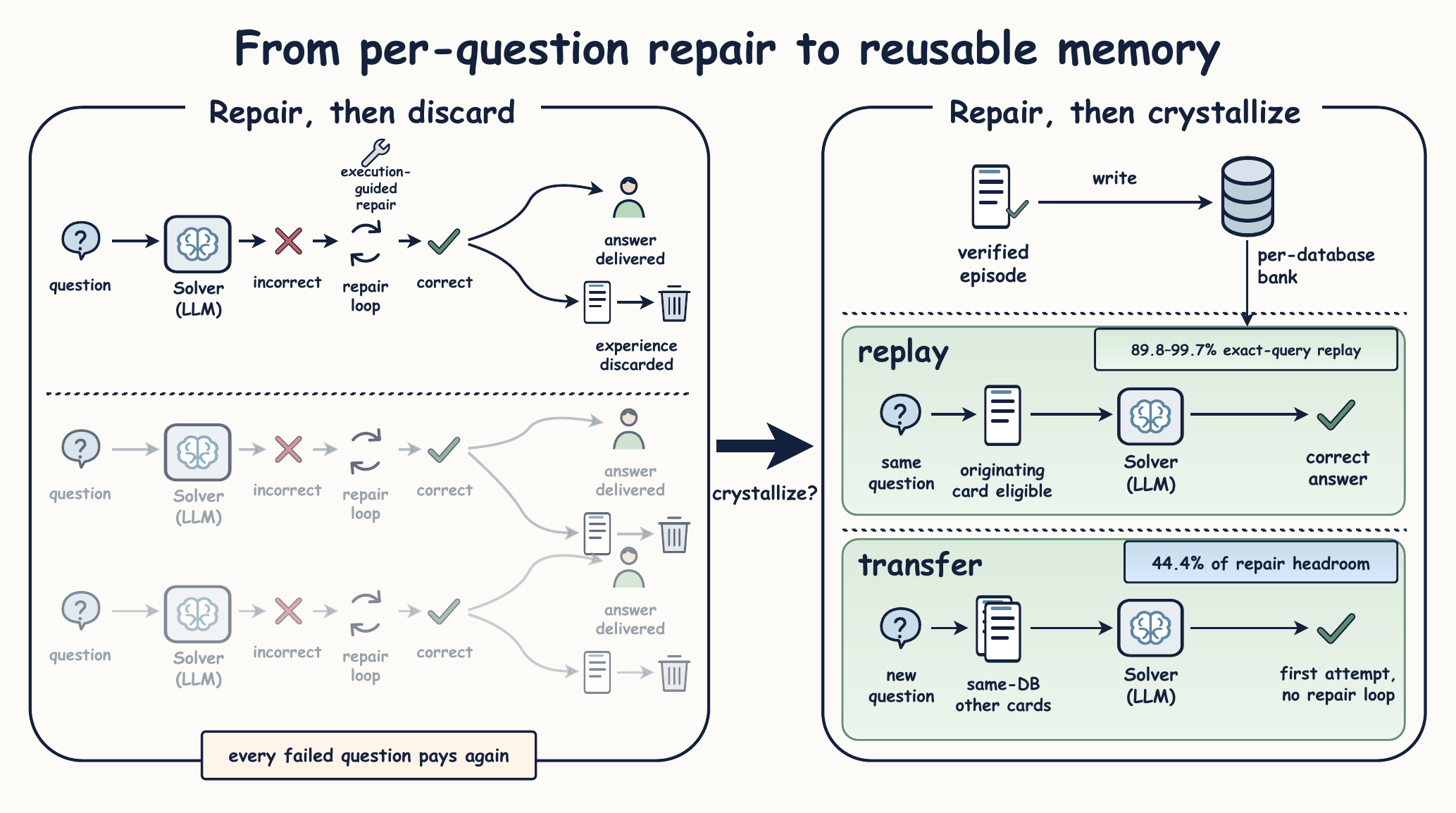}
\Description{Two side-by-side panels. Left, titled "Repair, then discard": a
question goes to a fixed LLM solver, the first attempt is incorrect, an
execution-guided repair loop spends extra computation, the corrected answer is
delivered to the user, and the episode's experience is discarded; faded repeats
show that every failed question pays this cost again. Right, titled "Repair, then
crystallize": the same verified episode is instead written into a per-database
bank, which then serves two routes: replay, answering the same question again
from the stored solution, and transfer, helping a new question over the same
database succeed on the first attempt without a repair loop. A badge reports
that the transfer lift is 44.4 percent of the on-demand repair headroom.}
\caption{\textbf{From per-question repair to reusable memory.} Left: discarding
a repaired episode makes each future failure repay the repair cost. Right:
verified episodes enter a database-scoped bank for exact-query \emph{replay}
and cross-question \emph{transfer}. On BIRD, transfer captures $44.4\%$ of the
held-out on-demand repair headroom (Section~\ref{sec:results}).}
\label{fig:teaser}
\end{figure}

%% file: sections/related.tex
\section{Related Work}
\label{sec:related}

\paragraph{Text-to-SQL with LLMs and execution-guided repair.}
LLM pipelines drive current progress on text-to-SQL
benchmarks~\cite{spider2018,bird2023} through question
decomposition~\cite{dinsql2023}, demonstration selection~\cite{dailsql2024},
schema linking~\cite{resdsql2023,chess2024}, and multi-candidate
selection~\cite{purple2024,deepeye2026}. The strongest models increasingly
relax the schema-linking step~\cite{deathschema2024}. Because a candidate
query can be run, a second line grounds refinement in the database itself:
multi-agent repair from execution errors~\cite{macsql2025}, self-debugging on
execution results~\cite{chen2024selfdebug,madaan2023selfrefine}, and
execution-grounded schema refinement~\cite{egrefine2026}. These methods are
sequential test-time scaling in the sense
of~\cite{snell2024tts,zhang2025ttssurvey}. Without an external signal,
intrinsic self-correction is unreliable~\cite{huang2024selfcorrect}. Common to
this line, compute is spent per question and the evidence a repair uncovers is
discarded once the query is emitted. Our study begins at that discard point.

\paragraph{Self-evolving text-to-SQL}
A fast-growing line keeps what inference discovers, in several forms: corrected
queries and worked examples~\cite{memosql2026,lpesql2024}, self-correction
guidelines and optimized hints~\cite{magic2025,tahoe2026}, semantic or dual-level
agent memories~\cite{agentsm2026,merit2026}, and distilled or curated domain
knowledge~\cite{orange2025,katsql2025,seed2025,knowledge2sql2024,tkboost2026,rubiksql2025}.
Other systems elicit knowledge continually from human feedback~\cite{continual2025}
or run autonomous evolution loops~\cite{robophd2026}. These systems accumulate
experience and report consistent end-to-end gains, and they directly motivate
our study. Their evaluations, however, score the loop as a bundle: gains on
recurrent questions can mix with gains on new ones, and the source, format, and
retrieval choices vary jointly with the surrounding pipeline. Several
mechanisms these systems rely on reappear as measured points on our source and
writer axes: self-consistency filtering~\cite{orange2025}, distilled
guidelines~\cite{magic2025}, and corrected-query
memories~\cite{memosql2026,lpesql2024}. We contrast mechanisms rather than
whole systems: a head-to-head score would entangle each system's pipeline,
prompts, and model with its memory design, which is precisely the bundling
this paper exists to undo. Table~\ref{tab:mechmap}
(Appendix~\ref{sec:x-audit}) maps each prior line to the cell that isolates
its mechanism.

\paragraph{Mechanistic studies of demonstrations.}
Our interventions follow mechanistic studies of in-context demonstrations,
where random labels preserve much of in-context
learning~\cite{min2022demonstrations} and invalid rationales retain most of
chain-of-thought performance~\cite{wang2023cotmatters}. Those studies concern
transient prompts; we test persistent cards built from the model's own earlier
episodes.

\paragraph{Memory and self-supervision beyond text-to-SQL}
Memory-augmented agents~\cite{yao2023react,sumers2024coala}
retain episodic insights, reflections, workflows, skills, strategies, and
test-time
cheatsheets~\cite{zhao2024expel,park2023genagents,shinn2023reflexion,wang2025awm,wang2024voyager,reasoningbank2025,suzgun2026cheatsheet},
or manage persistent memory as an explicit
resource~\cite{packer2023memgpt}. The same design questions recur:
retrieval can trail supervised training~\cite{exprag2026},
relevance-ranked recall admits misleading entries~\cite{li2026seam,cmi2026},
benchmarks separate retrieval, test-time learning, and forgetting as distinct
competencies~\cite{memoryagentbench2025}, and surveys caution that a single
aggregate score obscures how a memory
behaves~\cite{gao2026survey,shao2026misevolve,memtrace2026}. But the
trajectories are less structured, and attribution rests on task-specific
success proxies or LLM-based judgments; text-to-SQL pins the same questions to
deterministic, execution-based outcomes, which is what lets our answers take
the form of paired counts. On the supervision side, STaR bootstraps from
self-generated solutions kept by an answer check~\cite{zelikman2022star},
reinforcement learning turns execution feedback on gold-labeled data into
parametric
updates~\cite{sqlr1_2025,reasoningsql2025,cscsql2025,rewardsql2025,arctic2sql2025},
and agreement among samples serves as a confidence
signal~\cite{wang2023selfconsist}. Our setting is the non-parametric
counterpart under a deliberately thin interface: the memory writer sees the
model's episode, not the gold SQL; benchmark verification receives only
per-attempt correctness bits; every accepted write stays inspectable.
Parametric consolidation of the same episodes is outside the scope of
the present study (Section~\ref{sec:analysis}).

%% file: sections/framework.tex

\section{Study Formulation and Measurement Protocol}
\label{sec:framework}

\input{figures/fig2_protocol}

We organize the study around four choices. The source choice determines how an
episode is collected (\emph{acquisition}). The verification choice determines
whether it is stored (\emph{admission}). The format choice determines what the
stored card contains (\emph{writing}). The retrieval choice determines which
cards reach a future question (\emph{consumption}). The solver remains fixed,
and each comparison changes one choice. Figure~\ref{fig:protocol} summarizes
the protocol.

\subsection{Acquiring Test-Time Episodes}
\label{sec:framework-source}

A text-to-SQL instance is a question $q$ over a database $d$. A fixed solver
$S$ produces a first attempt $\hat{y}=S(q,d)$, and an execution check
$\mathrm{EX}(y,d)\in\{0,1\}$ compares a query's result set on $d$ against the
gold set. An \emph{experience source} $T$ converts the solver's test-time
computation into an episode $e = (q, \hat{y}, y^{+}, \tau)$: the question, first
attempt, accepted candidate, and interaction trace. Our main instantiation is
execution-guided repair of first-attempt failures under an attempt budget $K$.
The alternative sources test which feedback makes an episode reusable. They
differ in the signal $g$ available during generation. T1 receives none: it
resamples with zero feedback. T2 receives one bit: an oracle answers
``correct / not correct'' per retry. T3, the main source, receives
interaction: the model probes the database between attempts. T4 replaces
repair-time feedback with a self-vote: agreement among samples proposes a
candidate, and the oracle still decides whether it is stored
(Section~\ref{sec:framework-deploy} removes that final check). For all verified
sources, the oracle reveals only whether a candidate is correct; it never
reveals a gold SQL string. For comparison, we also store first-try-correct
answers as $y^{+}{=}\hat{y}$ and $\tau{=}\varnothing$. This lets us test whether
episodes produced by repair transfer better than ordinary successful answers
(Section~\ref{sec:results}).

\subsection{Card Construction Without Gold SQL}
\label{sec:framework-writer}

A card formatter $W$ maps an episode to a memory card $c = W(e)$. The formats
differ in how much of the episode they keep and in what form. The simpler
formats keep the episode nearly raw: the
question and corrected query, verbatim (W0); the entire repair trajectory,
unprocessed (W1); the edit as a structured diff over the query's syntax tree
(W2). The upper rungs add interpretation: the diff annotated with its
failure-mode label (W3), optionally paired with the corrected query (W3+A);
the annotated diff plus a guard condition stating when it applies (W4); a
distilled natural-language summary (W5). Card content and card verification are
separate. The writer receives only the episode, not the gold query; therefore,
the corrected query $y^{+}$ is the model's own output. We call this
\emph{gold-SQL-free writing}. A separate verification signal decides whether
the episode is stored (Section~\ref{sec:framework-deploy}). In benchmark mode,
that signal checks the execution result against gold. A sentinel test confirms
the separation: a marker placed in the gold-SQL field never appears in a card
or in the prompt shown to the solver.

\subsection{Database-Scoped Retrieval and Evaluation}
\label{sec:framework-consumer}

Each database has its own memory bank. It contains cards from candidate
episodes $\mathcal{E}_d$ that the verification signal $v$ accepts,
\begin{equation}
  M_d \;=\; \bigl\{\, W(e) \;:\; e=(q,\hat{y},y^{+},\tau)\in\mathcal{E}_d,\ v(y^{+})=1 \,\bigr\}.
  \label{eq:bank}
\end{equation}
The writer still never sees the gold query, although benchmark verification
$v$ uses the gold execution result (Section~\ref{sec:framework-deploy}).
Retrieval makes three choices: the matching key, the number of cards $k$, and
the database scope. By default, it matches the new question $q'$ against each
card's original question by cosine similarity and retrieves the top $k$ cards
from the same database. The stored SQL, diff, or summary does not affect this
match. Keeping the matching rule fixed lets us compare card formats without
giving any format a retrieval advantage. The solver then answers in one shot.
Section~\ref{sec:q3} varies retrieval width and database scope.
Every configuration is then evaluated in two strictly separated settings.
\emph{Transfer} is first-attempt accuracy on held-out questions from the same
databases. These questions were never used to build the bank, but they share
schemas and value spaces with the collection questions. \emph{Replay} and
\emph{retention} instead revisit collection questions that produced cards.
Exact-query replay allows the question's own card. Cross-question retention
removes that card and uses only other cards. The own-question card contains the
model's corrected query, never the gold SQL. Replay therefore measures a real
recurrence case rather than held-out generalization, and we report it
separately from transfer.

Algorithm~\ref{alg:r2m} (Appendix~\ref{sec:x-protocol}) summarizes the
end-to-end procedure.

\subsection{Measurement: Transfer, CR, and Cost}
\label{sec:framework-gauge}

Every quantity is an execution-accuracy mean over a fixed split. On a held-out
set $Q_{\mathrm{tr}}$ of $(q',d')$ pairs disjoint from the banked questions,
\emph{transfer} conditions the solver on the retrieved cards,
\begin{equation}
  P_M \;=\; \frac{1}{|Q_{\mathrm{tr}}|}\!\!\sum_{(q',d')\in Q_{\mathrm{tr}}}\!\!
    \mathrm{EX}\!\bigl(S\bigl(q';\,\rho_k(q';M_{d'})\bigr),\,d'\bigr),
  \label{eq:pm}
\end{equation}
Here, $P_0$ is accuracy without memory, and $P_K$ is accuracy after an
oracle-triggered repair pass on the same held-out questions. The repair pass is
used only for measurement; its episodes are never stored. The crystallization
ratio is
\begin{equation}
  \mathrm{CR} \;=\; \frac{P_M - P_0}{\,P_K - P_0\,}
  \label{eq:cr}
\end{equation}
It is the held-out first-attempt gain from memory, divided by the repair headroom on
the same questions. It does not claim that memory literally recovers the
collection episodes. CR is descriptive, not a probability: it can be negative
when memory is harmful or exceed $100\%$ when memory beats the measured repair
reference. We compare CR only when the solver, split, repair budget, and
collection procedure are the same. We also report the absolute accuracy gain
beside every headline CR (Appendix~\ref{sec:x-threats}).
\emph{Replay/retention} re-answers the once-repaired questions $Q_{\mathrm{bk}}$
themselves,
\begin{equation}
  R \;=\; \frac{1}{|Q_{\mathrm{bk}}|}\!\!\sum_{(q,d)\in Q_{\mathrm{bk}}}\!\!
    \mathrm{EX}\!\bigl(S\bigl(q;\,\rho_k(q;M_d)\bigr),\,d\bigr),
  \label{eq:ret}
\end{equation}
and we evaluate Eq.~\ref{eq:ret} both
with and without the originating card. The bare re-run floor
($0.8$--$3.1\%$ per seed; Appendix~\ref{sec:x-extended}) is reported but not
subtracted.
Finally, we report a per-question call count and an amortized lifecycle token
account (Section~\ref{sec:q4}; measured costs in Appendix~\ref{sec:x-deep});
neither is a complete dollar or latency model.

Table~\ref{tab:instantiation} (Appendix~\ref{sec:x-audit}) summarizes the
controlled factors, their defaults, and the interventions used to identify
their contribution. Section~\ref{sec:results} first measures reuse, then tests
what carries it, and finally compares the design choices that materially
affect it. Cost and cross-section probes are reported as supporting analyses.

\subsection{Verification Regimes}
\label{sec:framework-deploy}

The study requires a verification signal, and the experiments distinguish
three regimes. In \emph{benchmark verification}, used for the main measurements,
an oracle supplies per-attempt correctness bits and defines the repair ceiling
$P_K$. The gold SQL is not written, but its execution result determines
whether the episode is stored. In deployment, the same role could be played by a user confirmation,
an executable assertion, or a downstream business check. Our experiments
measure how reliable the signal must be (Section~\ref{sec:q3}), not which source
provides it. Finally, \emph{ungated self-vote storage} removes external
verification: the model stores the vote-elected candidate whether or not it is
correct.

The ungated bank performs worse than no memory in every seed
(Section~\ref{sec:q3}). Reliable verification is therefore necessary for the
measured benefit.

%% file: figures/fig2_protocol.tex
\begin{figure*}[t]
\centering
\includegraphics[width=0.96\textwidth]{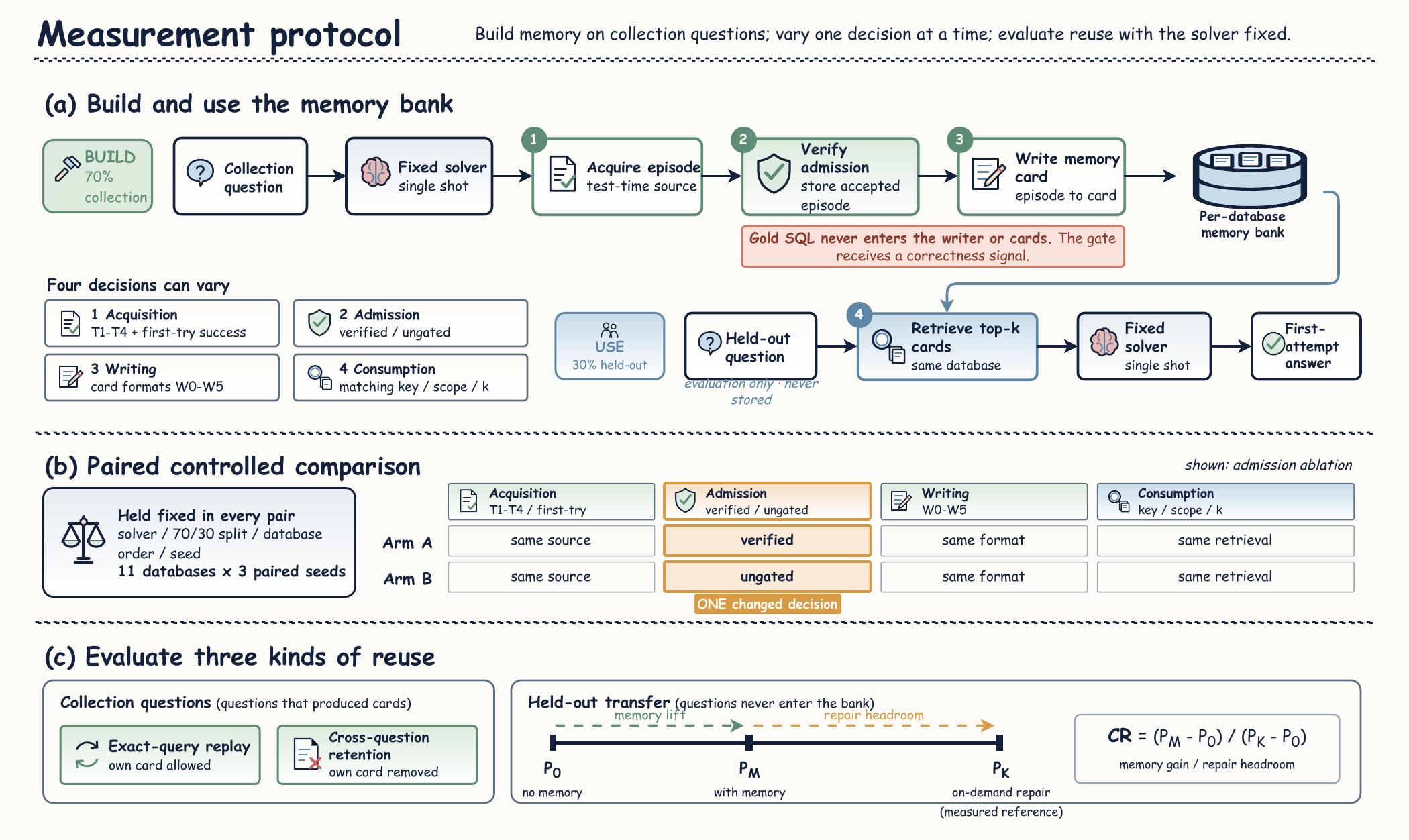}
\Description{Three stacked panels. Panel (a) shows collection questions passing
through a fixed single-shot solver, episode acquisition, admission, and
gold-SQL-free card writing into a per-database memory bank. Held-out questions
retrieve top-k cards from that bank and pass through the same fixed solver, but
are never stored. Four cards list the acquisition, admission, writing, and
consumption choices. Panel (b) shows a paired comparison that holds the solver,
split, database order, and seed fixed while changing only admission, illustrated
by verified versus ungated admission over eleven databases and three paired
seeds. Panel (c) shows exact-query replay, cross-question retention, and
held-out transfer at no-memory, memory, and on-demand-repair operating points,
together with the crystallization-ratio equation.}
\caption{\textbf{The measurement protocol.}
(a) We build per-database memory only from the $70\%$ collection split; the
$30\%$ held-out questions retrieve same-database cards but are never stored.
The four numbered steps mark acquisition, admission, writing, and consumption.
(b) Paired comparisons change one decision while holding the solver, split,
database order, and seed fixed within each pair; results aggregate 11 databases
and three paired seeds.
(c) Collection questions measure exact-query replay and cross-question
retention; held-out questions measure transfer at $P_0$ (no memory), $P_M$
(memory), and $P_K$ (on-demand repair). CR normalizes the memory gain by the
measured repair headroom (Eq.~\ref{eq:cr}).}
\label{fig:protocol}
\end{figure*}

%% file: sections/setup.tex
\section{Instantiation and Experimental Setup}
\label{sec:setup}

This section fixes the default value of every controlled factor;
Section~\ref{sec:results} varies one choice at a time against it.

\paragraph{Data and splits.}
Our main testbed is the BIRD development set~\cite{bird2023}: $1{,}534$ questions
over $11$ databases. For each seed ($42/7/13$) we draw a fixed $70/30$ split
\emph{within each database}: $1{,}073$ \emph{collection} questions, on which
repair runs and memory is built, and $461$ \emph{held-out} questions for transfer.
Because the split is per database, every database appears on both sides, so a
held-out question is a \emph{new question over a database the bank already
covers}. Throughout, ``transfer'' therefore means \emph{same-database} transfer:
serving a database's own future queries from memory built on it, a common
deployment setting. Cross-database generalization to unseen schemas is a
different deployment setting, probed directly in Section~\ref{sec:q3}. The two
sides are disjoint by construction (Section~\ref{sec:threats}). The second
benchmark is the Spider development set~\cite{spider2018} ($1{,}034$ questions;
$723/311$), run through the identical pipeline.
The collection cohort supplies the bank; its replay and retention
measurements are conditional on the verified-repaired questions that entered
it, and are reported separately from held-out transfer
(Section~\ref{sec:framework-consumer}).

\paragraph{Models.}
The main model is Qwen3.5-27B, with the full pipeline replicated on Qwen3.5-9B;
Gemma4-E4B, Gemma4-31B, gpt-oss-20b, and Hunyuan-A13B cover four families
(Section~\ref{sec:q5}). All models are served locally and decoded greedily
($T{=}0$, $2{,}048$-token budget, reasoning disabled), identically across arms.
Sampling occurs only inside the resampling sources (T1 one candidate, T4 five,
at $T{=}0.8$).
Batched serving keeps greedy decoding only approximately reproducible, so every
claim rests on within-chain paired contrasts. All prompts, card templates, and
a worked example are transcribed in Appendix~\ref{sec:x-protocol}.

\paragraph{Solver context.}
The solver sees the \emph{full} schema, built identically for every arm:
faithful CREATE-TABLE DDL plus declared foreign keys, with no sampled values.
We run no lossy schema-linking step~\cite{c3sql2023,rslsql2024,esql2024}, so
the retrieved memory block is the only content that varies across arms
(attribution by construction). The base solver is a single greedy call with no
multi-candidate decoding ($\sim$$62\%$ one-shot, $\sim$$72\%$ after three
repair rounds). Appendix~\ref{sec:x-threats} discusses how absolute levels
couple to this base.

\paragraph{Episode sources.}
For collection questions the solver gets wrong, each source may spend an
attempt budget $K{=}3$ to produce a candidate episode, using the feedback
settings in Section~\ref{sec:framework-source}. T3 (main) interleaves read-only
exploratory probes with retries; T2 retries on the correctness bit alone; T1
resamples with zero feedback. T4 samples five candidates per attempt,
executes them, clusters by result set, and accepts the majority. No
correctness signal enters this selection;
the oracle bit only controls storage in the verified arm. An otherwise identical
\emph{vote-admit} arm removes this final check and stores the first vote-elected
candidate. The origin contrast additionally banks first-try-correct solves and
their union with repairs. To separate episode origin from bank size, we downsample
the success bank to exactly the corresponding repair-bank count
($124$--$129$ cards per seed, pooled across that seed's databases). The draw is deterministic, made independently
twice, per database, preserving file order.

\paragraph{Card formats.}
We compare formats W0--W5, defined in Section~\ref{sec:framework-writer}.
We refer to them as \emph{Verbatim} (W0), \emph{Trace} (W1),
\emph{Diff} (W2), \emph{Diff+Mode} (W3), \emph{Anchor} (W3+A),
\emph{Diff+Guard} (W4), and \emph{Capsule} (W5). Diffs and guards are
deterministic (syntax-tree diff; fixed guard templates); only W5 capsules are
written by an LLM. An automatic check rejects capsules referencing
tables, columns, or values absent from the episode. Section~\ref{sec:q3}
compares the formats; the complete grid appears in
Appendix~\ref{sec:x-deep}. No format receives the gold SQL; an automated test
checks this boundary (Section~\ref{sec:framework-writer}).

\paragraph{Retrieval and evaluation.}
Cards are banked per database. At inference, the top-$k$ cards (same database,
$k{=}5$ unless stated) are retrieved by question-to-question cosine
similarity~\cite{reimers2019sbert} (Qwen3-Embedding-0.6B, $1{,}024$-d,
L2-normalized) and placed in the solver's context. The solver answers in one
shot. The correspondence intervention is specified where it is used
(Section~\ref{sec:q2}). We report execution accuracy (EX): correct iff the
result set matches the gold query's; unevaluable counts as wrong. Transfer, exact-query replay, and
cross-question retention follow the definitions of
Section~\ref{sec:framework-consumer}, scored by EX on their respective splits.
CR follows Eq.~\ref{eq:cr} with numerator and denominator on the held-out
split; the denominator is a measurement-only oracle-repair pass
($9.5$--$10.0$pp per seed) whose trajectories are never written to memory.

\paragraph{Statistical protocol.}
Every comparison changes one decision against the fixed default (T3, W0,
$k{=}5$) and uses three paired seeds. Our main analysis is a two-stage
hierarchical bootstrap. It first resamples the $11$ databases and then the
unique questions within each selected database; all seed results for a
question stay together. Four comparisons support the paper's main claims: W0
versus no memory, verified versus unverified storage, $k{=}10$ versus $k{=}1$,
and the question--SQL correspondence test. In the last test, permuted versus no
memory measures whether exact pairing is necessary, while aligned versus
permuted measures the additional value of correct pairing. We report two-sided
$p$ values and $95\%$ intervals.

Exact McNemar tests over question--seed pairs are supplementary diagnostics;
we do not treat those pairs as independent samples. We test equivalence with a
hierarchical TOST whose margin comes from observed rerun variability. No card
format or episode-source comparison passes that test. The remaining source,
format, retrieval, model, and benchmark sweeps are exploratory. We treat their
$p$ values as diagnostics, and report experiments with fewer than three seeds
as preliminary (Section~\ref{sec:threats}). Deltas and ratios are computed
before rounding.

%% file: sections/results.tex

\section{Results}
\label{sec:results}

The first three subsections answer RQ1--RQ3: how much value memory provides,
what information carries transfer, and which memory choices matter. We then
report cost and robustness. Unless stated otherwise, experiments use BIRD,
Qwen3.5-27B, T3 episodes, W0 cards, $k{=}5$, three paired seeds, and the fixed
solver of Section~\ref{sec:setup}.

\subsection{What Future Value Does the Bank Provide?}
\label{sec:q1}

\input{figures/fig2_crystallization}
\input{tables/table2_cr_main}

Table~\ref{tab:cr-main} reports the collection-side and held-out settings
separately. On held-out questions, Figure~\ref{fig:crystallization} shows that
verbatim corrected queries lift one-shot accuracy from $62.04\%$ to $66.38\%$.
The two-stage bootstrap gives a gain of $+4.34$pp (CI $[+1.50,+7.49]$,
$p{=}.0034$). The corresponding $\mathrm{CR}=44.4\%$ means that memory captures
$44.4\%$ of the on-demand repair headroom on the same held-out questions. The
ratio has a wider interval of $[24,65]\%$.
Ten of eleven databases improve. Memory fixes 127 held-out questions and
breaks 67, for a net gain of $60$ among $1{,}383$ question--seed cases. The
gain also survives the stricter subset metric in every seed ($+3.9$ to
$+5.2$pp). Across seeds, 25 questions are stably fixed and five are stably
broken; no question changes from a fix to a break
(Appendix~\ref{sec:x-deep}). On the
collection questions that produced cards, exact-query replay reaches $96.1\%$
when the question's own card is available. Cross-question retention falls to
$56.0\%$ when that card is removed. Section~\ref{sec:analysis} explains why
these settings must remain separate.

\subsection{What Information Carries Cross-Question Transfer?}
\label{sec:q2}

\input{figures/fig3_mechanism}

We test three explanations. First, any well-formed Q--SQL examples may help.
Second, the model may need a correctly matched example from a similar question.
Third, cards may help by exposing identifiers, values, joins, and query
structures from the target database, even when the example pairings are
imperfect.

Figure~\ref{fig:mechanism} summarizes these tests; Table~\ref{tab:rq2-mechanism}
(Appendix~\ref{sec:x-audit}) gives the statistics. Local cards beat equally
valid foreign cards by $6.73$pp. Foreign cards have a negative point estimate
against no memory, but that separate confidence interval includes zero after
database clustering. Our conclusion therefore rests on the paired
local--foreign comparison, not on a claim that every foreign bank is harmful.
Reducing local cards to natural-language lessons costs $2.68$pp. The stored SQL
and database details therefore add useful information.

Targeting adds a distinct contribution. Similarity-targeted local cards beat
same-database random retrieval by $2.03$pp, whereas random-local versus no
memory has a positive point estimate but a confidence interval that includes
zero. Question similarity therefore helps select relevant content, but this
result does not show that the model copies a matched example.

The correspondence intervention tests that stronger imitation account. We
permute corrected SQLs within each database while preserving the question and
SQL multisets, card count, token distribution, retrieval order and scores, and
prompt format. Within this paired rerun, the aligned bank improves accuracy by
$5.36$pp and the mispaired bank by $3.91$pp. Mispairing therefore preserves
$73\%$ of the aligned lift. The aligned estimate is $1.02$pp above the headline
estimate, consistent with measured rerun variability; we use the paired
aligned--permuted difference as the evidence about correspondence. Restoring
correct pairing adds $1.45$pp, but its confidence interval includes zero (CI
$[-0.45,+3.23]$, $p{=}.14$). Correct mappings may add a smaller benefit, but
they are not required for most of the observed transfer. Together, these tests
show that the cards mainly provide database-specific information rather than
matched examples to copy.

\subsection{Which Experience-Memory Choices Materially Affect Transfer?}
\label{sec:q3}

\input{figures/fig4_decisions}

Figure~\ref{fig:decisions} shows the controlled effects.
Table~\ref{tab:rq3-decisions} (Appendix~\ref{sec:x-audit}) reports which
comparisons are statistically supported and which remain inconclusive.

\emph{Episode source.} Banks built from first-try successes and repaired
episodes have similar point estimates. At matched per-database counts, repaired
cards are higher by $1.30/2.32$pp in two independent draws, but both confidence
intervals include zero. Repair efficiently focuses storage on failures and
creates cards for replay, but the data do not show that repair episodes are
uniquely useful for transfer. The source still matters: zero-feedback
resampling and one-bit retry underperform interactive, probe-grounded repair at
matched bank sizes, while pre-answer exploration alone reaches only about
$37\%$ of the on-demand repair headroom on collection questions.

\emph{Verification.} This choice has the clearest effect. Verified self-vote
cards beat their otherwise matched unverified bank by $4.85$pp (CI
$[+1.30,+8.06]$, $p{=}.0064$); the ungated bank is $2.03$pp \emph{below} no
memory, negative in all three seeds. Only $3.9\%$ of raw vote selections pass
the oracle, and the audited ungated bank is $5.6\%$ correct. The gate need not
be perfect, however: synthetically flipping $5$--$20\%$ of verification decisions
leaves transfer above the no-memory floor at every level
(Appendix~\ref{sec:x-deep}). Agreement can propose candidates, but it cannot
replace reliable verification before storage.

\emph{Card format.} Verbatim W0 has the highest transfer point estimate; every
richer writer is lower by $1.3$--$2.0$pp. The largest deficit is the unprocessed
full trace, but all intervals cross zero. None passes hierarchical equivalence
even at a $\pm2.5$pp margin. Thus, richer formats do not show an improvement,
but the data also do not establish that all formats are equivalent. Format
matters more for exact replay: mean accuracy across seeds ranges from
$89.8\%$ to $99.7\%$ when the question's own card is available. After removing
that card, W0 and W5 give similar retention ($56.0/56.8\%$). Richer formatting
therefore helps reproduce some stored answers, but shows no clear advantage for
cross-question retention or held-out transfer (Appendix~\ref{sec:x-extended}).

\emph{Retrieval.} Increasing the number of retrieved cards from one to ten adds
$3.18$pp (CI $[+1.09,+5.35]$, $p{=}.0028$). The individual steps between these
endpoints are not statistically significant, so we claim only the overall
$k{=}1$ to $k{=}10$ gain. At $k{=}10$, verbatim reaches
$\mathrm{CR}=57.6\%$. Retrieval width and bank size interact: using half of the
bank matches the full bank, while using only a quarter causes accuracy to fall.
A wider retrieval window helps only when enough cards are available.

Once the bank has enough cards and retrieval stays within the target database,
the retrieval algorithm matters less. The four targeted methods differ by less
than $0.7$pp; BM25 matches dense embeddings, and a payload-aware key does not
change the ordering of card formats.

\subsection{Supporting Analyses: Cost, Robustness, and Scope}
\label{sec:q4}

\emph{Cost.} We measure cost on a separate, internally paired serving run
(Table~\ref{tab:amortization}, Appendix~\ref{sec:x-extended}). In that run,
memory lifts the first attempt by
$+2.7$pp, and one repair round captures $80\%$ of what
three memory-less rounds achieve at $59\%$ of the calls. Memory does not raise
the repaired endpoint ($70.0$ vs.\ $70.4$), so the two gains mainly substitute.
At question level, memory-less repair resolves $59$--$67\%$ (mean $62\%$) of
the questions memory fixes, versus $26\%$ of comparable base failures. This
$2.4\times$ enrichment supports substitution rather than an independent
endpoint gain (Appendix~\ref{sec:x-deep}).
Building a T3 bank costs $5.90$--$6.04$M tokens per seed; serving adds
$139/604/1{,}092$ prompt tokens at $k{=}1/5/10$. Relative to the measured
serve-time repair reference, W0 uses $2.6\times$ fewer recurring prompt tokens
per lift point. This token account reaches its amortization crossover after
roughly 7.5--9.9K future queries. It is not a dollar,
latency, energy, or total-cost-of-ownership estimate: embedding and database
execution remain in their native units (Appendix~\ref{sec:x-threats}).
A second pass collapses repair yield ($\sim$$31\%\to9$--$13\%$) and moves
transfer by $-0.7$pp on average, making the first collection pass the efficient
operating point (Appendix~\ref{sec:x-deep}).

\emph{Robustness.}\label{sec:q5}
At 9B, memory adds $+6.2$pp (CI $[+4.0,+8.5]$), equal to $49.5\%$ of the
on-demand repair headroom. This replicates the benefit at a second Qwen scale,
but does not show that the benefit increases with model size.
Ten of eleven BIRD databases gain, although bank size barely predicts lift
($r{=}.18$). Single-seed tests across four model families have positive point
estimates ($+3.3$ to $+6.9$pp). They do not include the held-out repair pass
needed to calculate CR. On Spider, replay ordering repeats while transfer deltas
are small, consistent with the mechanism: Spider's no-memory floor is already
$79$--$80\%$, leaving little repair headroom to crystallize, and its questions
require less database-value grounding than BIRD's~\cite{bird2023}. We treat
these tests as preliminary evidence about scope, not as full replications
(Appendix~\ref{sec:x-extended}).
Memory also composes with a chain-of-thought base, adding $+2.5$pp on 27B
($p{=}.022$) and $+4.8$pp on 9B ($p{=}2.6\text{e-}5$). The smaller lift is
consistent with measured redundancy: chain-of-thought already solves $43\%$
of memory-fixed questions (Appendix~\ref{sec:x-deep}).

%% file: figures/fig2_crystallization.tex

\begin{figure*}[t]
\centering
\includegraphics[width=0.75\textwidth]{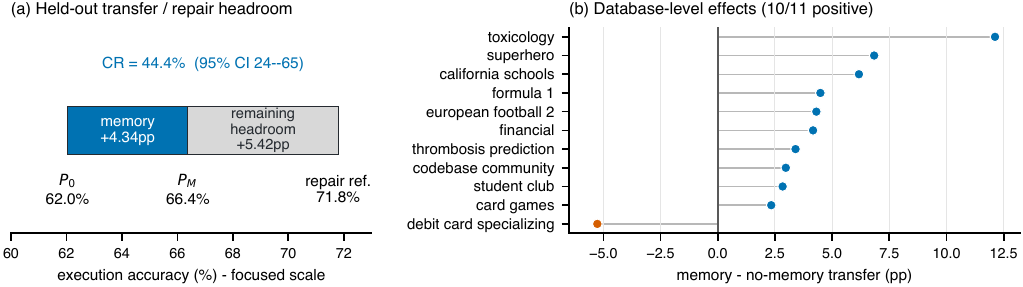}
\caption{\textbf{RQ1: held-out transfer reaches a substantial share of
on-demand repair headroom.} W0 raises held-out same-database first-attempt
accuracy by $4.34$pp, yielding $\mathrm{CR}=44.4\%$ (95\% CI $[24,65]\%$):
the transfer lift is $44.4\%$ of the repair headroom measured on those same
held-out questions. Panel (b) shows positive effects in 10 of 11 databases.
The focused accuracy scale in panel (a) is marked explicitly.}
\Description{Panel a compares a 4.34-point held-out transfer lift with
held-out on-demand repair headroom.
Panel b shows positive transfer effects for ten of eleven databases.}
\label{fig:crystallization}
\end{figure*}

%% file: tables/table2_cr_main.tex
\begin{table}[t]
\caption{Complementary future-value settings that a single memory score would
conflate (BIRD, Qwen3.5-27B, three seeds). Replay and retention re-answer the
verified-repaired bank questions (retention excludes the originating card);
transfer uses 461 held-out questions per seed. CR normalizes the transfer lift
by held-out repair headroom (Eq.~\ref{eq:cr}); it is not a collection-repair
recovery rate. Per-seed grids: Appendix~\ref{sec:x-extended}.}
\label{tab:cr-main}
\centering\footnotesize
\setlength{\tabcolsep}{3.6pt}
\resizebox{\columnwidth}{!}{%
\begin{tabular}{@{}lccc@{}}
\toprule
Setting and quantity & Reference & W0 memory & Summary \\
\midrule
\multicolumn{4}{@{}l}{\emph{Collection-side: conditional on verified-repaired bank questions}} \\
Exact-query replay & 1.8\% & 96.1\% & originating card eligible \\
Cross-question retention & 1.8\% & 56.0\% & originating card excluded \\
\addlinespace
\multicolumn{4}{@{}l}{\emph{Held-out: new same-database questions}} \\
Same-DB transfer EX & 62.04\% & 66.38\% & $+4.34$pp $[+1.50,+7.49]$ \\
CR (transfer / repair headroom) & --- & 44.4\% & 95\% CI $[24,65]\%$ \\
\bottomrule
\end{tabular}
}
\end{table}

%% file: figures/fig3_mechanism.tex

\begin{figure*}[t]
\centering
\includegraphics[width=0.70\textwidth]{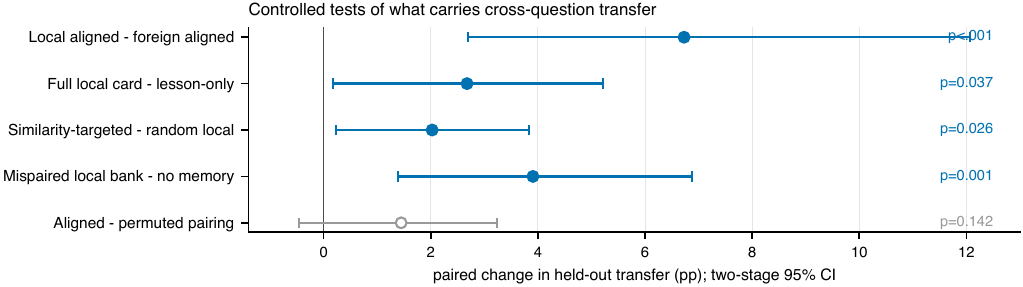}
\caption{\textbf{RQ2: database-specific content is the main operating ingredient for observed transfer.}
Local cards, grounded content, and similarity-based selection all help. A
fully mispaired local bank remains above no memory, while restoring the correct
question--SQL pairing adds $1.45$pp with a confidence interval that includes
zero. Bars are two-stage 95\% intervals; filled points pass $p<.05$.}
\Description{A five-row interval plot shows significant positive effects for
locality, grounding, targeting, and a mispaired local bank; the incremental
effect of restoring exact question-SQL correspondence has an interval that
includes zero.}
\label{fig:mechanism}
\end{figure*}

%% file: figures/fig4_decisions.tex

\begin{figure*}[t]
\centering
\includegraphics[width=0.70\textwidth]{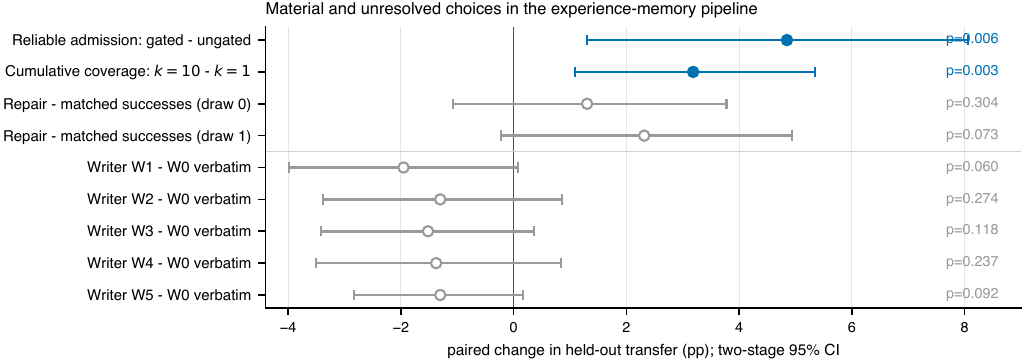}
\caption{\textbf{RQ3: reliable verification and wider retrieval improve
transfer.} Verification adds $4.85$pp, and retrieving ten cards adds $3.18$pp
over retrieving one. The intervals for repair-specific episodes and richer
card formats include zero; filled points pass $p<.05$.}
\Description{A nine-row interval plot shows significant gains from verified
verification and wider retrieval, while the source and card-format contrasts
have intervals that include zero.}
\label{fig:decisions}
\end{figure*}

%% file: sections/analysis.tex
\section{Discussion}
\label{sec:analysis}

\paragraph{What the controls show.}
No single intervention isolates every component of a memory card, but the
controls show that two broad explanations cannot, on their own, account
for the main effect
(Section~\ref{sec:q2}). Generic examples alone are insufficient: correctly
matched cards from the same database beat equally valid cards from other
databases. Copying a worked example is not the main explanation either:
breaking every local question--SQL match preserves most of the gain. Together,
these controls indicate that cards help mainly by exposing the solver to useful
database-specific content (with limits in Appendix~\ref{sec:x-threats}).
The retrieved windows are broad rather than template-dominated: $72$--$73\%$
of cards appear in at least one fixed question's window, while the top decile
accounts for only $27$--$32\%$ of appearances. This is distributional rather
than card-level causal evidence, but it supports the database-payload reading
(Appendix~\ref{sec:x-deep}).
The controls locate an operating ingredient, not causal shares: they do not
separate identifiers, values, joins, SQL distribution, and generic prompt
format. The small positive residual for correct question--SQL pairing remains
unresolved under the hierarchical test. In an upper-bound verbatim trace, a
card element appears in $77\%$ of changed fixes and $75\%$ of changed breaks.
The memory tax is therefore consistent with the same content channel reaching
questions that did not need it, rather than a separate failure mode
(Appendix~\ref{sec:x-deep}).

\paragraph{Transfer and replay are complementary future-use settings.}
Exact-query replay rewards stored solutions when questions recur. By contrast,
cross-question retention is about $56\%$, while held-out transfer is more
sensitive to scope, targeting, and cumulative coverage. A single score would
obscure whether bank value comes from recurrence, re-derivation, or assistance
on new questions; separate reporting makes these routes visible.

\paragraph{Scope and open questions.}
The results favour a simple operating point, not a universal recipe. The
main evidence concerns same-database reuse on BIRD with a fixed
single-shot solver; CR replicates at two Qwen scales, while model-family cells
are single-seed probes and Spider's small transfer deltas track its collapsed
repair headroom (Section~\ref{sec:q5}). Near-duplicate schemas across tenants,
moving cards into model weights, forgetting less useful cards, and multi-agent solvers
remain untested: these bound generalization rather than the present controls.

\paragraph{Managing persistent memory.}
Two operational risks follow directly. Scope is a data-isolation boundary:
cards contain database-local values and identifiers, so cross-database
(in deployment, cross-tenant) retrieval is both a quality failure and a
potential leakage channel. Verification determines what is stored. The cards
remain easy to inspect and delete only when an independently reliable check
controls storage.

%% file: sections/threats.tex
\section{Threats to Validity}
\label{sec:threats}

\paragraph{Leakage.}
Repaired and held-out questions are disjoint per seed and model; ids are
deduplicated, and a sentinel test confirms that gold SQL never enters cards.
Near-duplicate and template recall are negligible: only $0.1\%$ of nearest
neighbours paraphrase a held-out question and $0.7$--$0.9\%$ have a same-template
card; excluding either group leaves the lift unchanged. Twelve identical-config
run pairs bound rerun drift at median $0.65$pp and maximum $1.74$pp
(Appendix~\ref{sec:x-deep}).

\paragraph{Statistical dependence and multiplicity.}
Questions cluster within eleven databases and recur across seed splits. The
two-stage bootstrap resamples databases, then unique questions while keeping
seed replicates together. The main lift remains positive ($+4.34$pp, CI
$[+1.50,+7.49]$, $p{=}.0034$; ten of eleven databases positive); CR has a wider
$[24,65]\%$ interval and is therefore reported beside the absolute lift.
We reserve confirmatory claims for the four comparisons identified in
Section~\ref{sec:setup}. Other sweeps and secondary $p$ values are exploratory.

\paragraph{Seeds and single-seed cells.}
Every main claim is paired over three seeds. Results with fewer than three
seeds are labeled preliminary and do not establish the study's conclusions.

\paragraph{Benchmark scope.}
Two development-set benchmarks cover one task family because hidden test sets
preclude repeated paired probing. Absolute levels may inherit development-set
optimism; paired contrasts reduce serving drift, not repeated-analysis bias.
Enterprise-scale settings such as ScienceBenchmark~\cite{sciencebench2024} and
Spider~2.0~\cite{spider2_2025} remain outside this study.

\paragraph{Deployment workload.}
The $70/30$ protocol measures conditional values in distinct future-use
settings, not their prevalence in a production workload. We therefore do not
combine replay, cross-question retention, and held-out transfer into one
expected-utility or break-even claim. Applying the results requires a target
workload's recurrence distribution and cost model.

\paragraph{Further threats.}
Appendix~\ref{sec:x-threats} adds analyses of metric conventions, oracle and
intervention limits, base coupling, recurrence, and the detailed workload and
cost scope.

%% file: sections/conclusion.tex
\section{Conclusion}
\label{sec:conclusion}

Using execution-guided repair as a controlled instance of test-time scaling,
we measured how verified episodes create future value for same-database
text-to-SQL. After separating exact-query replay, cross-question retention,
and held-out transfer, we find that transfer from a verbatim bank captures
$44\%$ of the available repair headroom. Controlled interventions support a
store of database-specific information built from past experience, rather than merely a
book of matched examples. Reliable verification and broad retrieval coverage
produce clear gains. Richer card formats, more elaborate retrievers, and
repair-specific episodes show no statistically supported advantage. By
separating these effects, the study shows where reusable memory creates value
and which choices materially affect it.

%% file: sections/x_extended.tex

\section{Per-Seed Results and Full Grids}
\label{sec:x-extended}
\suppressfloats[t]

This section expands the means reported in Section~\ref{sec:results} to their
per-seed values and full grids; every mean supporting a main claim has its
raw layer here.

\input{tables/table3_amortization}
\input{figures/fig4_consumption}
\input{tables/table4_crosssections}

\paragraph{Held-out repair headroom.}
Table~\ref{tab:x-pk} lists the measurement-only held-out oracle-repair headroom
behind every CR. The 27B gap is tight across seeds ($9.5$--$10.0$pp); the 9B gap spans
$11.7$--$14.5$pp, which is the primary driver of the 9B CR variance in
Table~\ref{tab:x-9b}.

\input{tables/xtabB0_pk}
\input{tables/xtabB1_writers}

\paragraph{Complete writer ladder.}
Table~\ref{tab:x-writers} reports the complete W0--W5 ladder using one
collection-side evaluation chain and writer-level means over the three seeds.
W0 has the strongest transfer point estimate, while every richer writer is
lower; under the two-stage analysis their intervals cross zero and no writer
passes hierarchical equivalence. The replay range is $89.8$--$99.7\%$ (W4 to
W1/W5), but this is exact-query replay: excluding the self-card collapses W0
and W5 to $56.0/56.8\%$ at $k{=}5$.

\input{tables/xtabB2_sources}

\paragraph{Sources, per seed.}
Table~\ref{tab:x-sources} adds the card counts that make the original source
panel an equal-card comparison: each matched-T3 subsample is drawn to its source's
exact yield (51--59 for T1/T2, 46--50 for T4). The verified T4 and matched-T3
point estimates remain close, but this does not justify storing cards without verification:
the end-to-end ungated T4 bank is $4.85$pp below the verified bank
(Table~\ref{tab:x-audit}).

\input{tables/xtabB3_9b}
\input{tables/xtabC1_kwriter}
\input{tables/xtabC2_banksize}

\paragraph{Retrieval grids.}
Tables~\ref{tab:x-kwriter} and~\ref{tab:x-banksize} give the full
writer$\times$width and bank$\times$width grids behind the cumulative-coverage
row of Figure~\ref{fig:decisions}. Every writer's mean column is ordered with $k$,
but individual seed columns are not (most visibly, seed 42 dips at $k{=}3$ for
all three writers). Under the two-stage analysis, the supported inferential
claim is the W0 cumulative $k{=}10-k{=}1$ gain; none of the intermediate steps
is individually statistically significant. In the bank grid,
the $k{=}10$ advantage at the full bank is positive in all three seeds
($+1.96/+0.65/+1.30$pp), while at quarter bank it is absent or negative.

\input{tables/xtabB4_controls}
\input{tables/xtabF2_permuted_content}

\paragraph{Retrieval controls, per seed.}
Table~\ref{tab:x-controls} expands the three controls of
Section~\ref{sec:q2}: both cross-database arms are below the floor in every seed
(the misleading-card effect is not an averaging artifact), and the
random-retrieval arm's gap to top-$k$ widens from $0.9$pp at seed 42 to
$3.3$pp at seed 13. The schema-similar arm retrieves from each database's
nearest neighbour by CREATE-DDL embedding cosine; BIRD's closest pair reaches
only $0.60$, so the near-duplicate regime stays out of reach here.
Table~\ref{tab:x-permuted} then isolates question--SQL correspondence while
holding the local bank and retrieval rule fixed. Correct pairing has a positive
point estimate in all three seeds ($+1.52/+0.87/+1.95$pp), but its hierarchical
confidence interval includes zero. The supported result is that the mispaired
local bank remains above no memory in ten of eleven databases. Random,
permuted, and aligned are separate controls for targeting and correspondence;
they are not a complete factorial design and cannot estimate an interaction.

\input{tables/xtabC5_srcaxis}
\input{tables/xtabF1_matched_admission}

\paragraph{Origin contrast, per seed.}
Table~\ref{tab:x-srcaxis} expands the episode-source analysis in
Section~\ref{sec:q3}, including the
run-internal no-memory floor used for pairing (it differs from the main floor by
up to $0.4$pp of the serving drift noted in Section~\ref{sec:setup}, which is
why deltas are computed within-run).
Table~\ref{tab:x-matched-admission} then removes the bank-size confound and
reports the end-to-end verification control that supersedes the earlier T4-only
interpretation.

\input{tables/xtabC3_families}
\input{tables/xtabC7_perdb}

\paragraph{Families and databases.}
Table~\ref{tab:x-families} gives the full family grid (including the anchored
writer where measured); Table~\ref{tab:x-perdb} gives the per-database
decomposition of the main transfer result, the raw layer behind the
heterogeneity paragraph of Section~\ref{sec:q5}.

\paragraph{Cost detail.}
The per-configuration call counts behind Table~\ref{tab:amortization} are
three-seed means from the cost chain. Its lifecycle panel additionally uses raw
collection logs: 5.96M full / 4.56M marginal construction tokens, 4.3K database
probes per seed, and measured $k$-dependent serving prompt overhead.

\section{Measurement Audit}
\label{sec:x-audit}


\paragraph{Claim audit.}
Table~\ref{tab:x-audit} lists every comparison supporting a main claim with its paired
evidence; no number in Section~\ref{sec:results} lacks a row here.

\input{tables/table1_instantiation}
\input{tables/table_rq2_mechanism}
\input{tables/table_rq3_decisions}
\input{tables/table_mechmap}
\input{tables/xtabD1_audit}

\paragraph{Convention sensitivity.}
Two accounting choices could move the headline and do not. Replacing the CR
reference headroom (oracle-pass gap) with the alternative
$P_K - \mathrm{EX}_{none}$ moves every CR by less than $2$pp and reorders
nothing. Scoring EX over
evaluable-only questions instead of all questions moves means by at most
$0.15$pp (one question); we use the stricter all-questions convention
throughout.

\paragraph{Hygiene checks.}
Per seed, model, and benchmark: repaired-question ids intersect held-out ids in
the empty set (verified programmatically after every collection, including the
resumed run, whose duplicated log lines were deduplicated by question id before
any count); a sentinel unit test verifies that the gold field cannot reach a
card or a rendered prompt; paired tests restrict to questions evaluable in both
arms.

\section{Extended Threats to Validity}
\label{sec:x-threats}

This section extends Section~\ref{sec:threats} with seven further threat
analyses.

\paragraph{Gauge and metric.}
CR is a held-out transfer lift normalized by the on-demand repair headroom on
the same held-out questions; it is not a literal recovery rate for collection
repairs. Its numerator and denominator must share a split and collection stack.
All reported CRs enforce this convention, and the alternative denominator moves
every CR by less than $2$pp (Appendix~\ref{sec:x-audit}). EX
can pass a semantically wrong query on a lucky database state; this affects all
arms equally under paired tests, and the headline survives the stricter subset
metric in every seed (Appendix~\ref{sec:x-deep}).

\paragraph{Oracle scope.}
Collection uses correctness bits to decide what enters a bank (benchmark mode,
Section~\ref{sec:framework-deploy}); the gold SQL never enters card content, but
the construction is not oracle-free. The external signal is necessary at the
verification step: ungated self-vote memory is $-2.03$pp below no memory and $-4.85$pp
below its otherwise matched verified counterpart (two-stage $p{=}.0064$). The
gate need not be perfect: synthetically flipping the verification decision on
$5$--$20\%$ of episodes leaves transfer above the no-memory floor at every
level (Appendix~\ref{sec:x-deep}). Synthetic label noise and endogenous self-vote
errors are not interchangeable, however: the latter are correlated, grounded
mistakes and are substantially more damaging. Deployment therefore requires an
independently reliable verification channel; we do not claim that the model can
safely decide what to store on its own.

\paragraph{Mechanism intervention.}
The body states the causal boundary; the correspondence intervention does not
support a strong imitation account:
the permuted bank retains $73\%$ of the aligned lift, while the residual aligned
advantage is $+1.45$pp, but its interval includes zero (CI
$[-0.45,+3.23]$pp, $p{=}.14$). We therefore reject the strong imitation
account rather than treating that point estimate as evidence. Combined with
the direct local--foreign contrast ($+6.73$pp,
$p{<}.001$) and grounding-stripping control ($+2.68$pp, $p{=}.037$), the
intervention indicates that useful database-specific content is the main explanation,
but it does not assign causal shares to schema identifiers, values, joins,
aggregate SQL distribution, or generic prompt-format effects.

\paragraph{Base pipeline.}
The base solver is a controlled single greedy call over the full schema
(Section~\ref{sec:setup}), so the memory block is the only moving variable;
absolute levels and possibly crystallization fractions are coupled to this
base. Two results bound that coupling: the memory lift persists over a
chain-of-thought base, shrinking with measured redundancy, and the three-seed
Qwen scale replication remains positive. Other model-family cells are
single-seed probes, not robustness guarantees (Section~\ref{sec:q5}).
Reproduction on a multi-agent solver remains open.

\paragraph{Recurring collection questions.}
Replay retrieval can surface the card built from the question itself, by design
(Section~\ref{sec:framework-consumer}). This is valuable if questions recur, but
it is a different deployment setting from held-out transfer. We therefore report
the decomposition in the main table and the complete writer ladder:
$89.8$--$99.7\%$ exact-query replay across writer-level three-seed means becomes
about $56\%$ for verbatim and capsule when the self-card is excluded. The latter
measures re-derivation from other episodes within the banked collection cohort.
Transfer never retrieves a self-card and supports the unseen-question claim
regardless.

\paragraph{Unmeasured workload mix.}
Section~\ref{sec:threats} states the main boundary: the $70/30$ protocol gives
conditional values, not the prevalence of recurrence, cross-question reuse, or
entirely new questions in a production workload. A target-workload estimate
therefore additionally needs its recurrence distribution and cost model; the
conditional quantities here cannot be combined into one expected utility.

\paragraph{Cost scope.}
The body records the scope of the lifecycle account. Its measured inputs are
LLM tokens, calls, prompt overhead, and database probe counts. The 7.5--9.9K-
query horizon equates construction tokens with cumulative prompt-token overhead;
embedding and database execution remain in native units rather than being
assigned arbitrary exchange rates.

\section{Robustness, Cost, and Retrieval Details}
\label{sec:x-deep}

\input{tables/xtabE1_tokens}
\input{tables/xtabE2_difficulty}

\paragraph{The rerun-drift band, defined.}
Several comparisons in the body are called ``within rerun drift''; this band is
measured, not assumed. Across twelve pairs of runs with identical configuration
(same frozen bank, split, writer, $k$, greedy decoding) re-executed on different
days, nodes, or serving pools, the absolute difference in pooled EX has median
$0.65$pp and maximum $1.74$pp, and $1.5$--$2.4\%$ of individual answers flip
between two identical-config runs: greedy decoding is not bit-stable across
serving environments. A same-config contrast below $\sim$$1.7$pp therefore falls within the
observed rerun variation; the retrieval-method cluster of
Section~\ref{sec:q3} ($0.7$pp spread) sits well inside that band.

\paragraph{Metric and difficulty robustness.}
The body reports that the headline survives the stricter subset metric in every
seed. Table~\ref{tab:x-difficulty} gives the full audit: subset matching retains
a $+3.9$ to $+5.2$pp lift in the three seeds (pooled fix/break $123/61$,
$p{=}6\times10^{-6}$), including the seed that is borderline under EX.
Stratified by BIRD's difficulty labels, all three strata gain significantly,
with the largest lift on \emph{moderate} questions ($+6.4$pp) and the smallest
on \emph{simple} ($+3.0$pp), consistent with the payload reading: simple
questions rarely miss a bankable local fact, while moderate ones fail on
exactly the value and join conventions cards carry.

\paragraph{Paired tests for the ladder and retriever cells.}
Naive pooled McNemar tests make several writer contrasts appear significant,
but the two-stage analysis changes the reading: all W1--W5 intervals against W0
cross zero, W1 has the lowest point estimate ($-1.95$pp, $p{=}.065$), and none passes
hierarchical equivalence. The supported conclusion is therefore ``W0 has the
best point estimate; no richer writer shows a reliable gain.'' On retrieval
method, BM25 versus dense is a paired null ($p{=}.80$), as are MMR versus
dense ($p{=}.22$) and MMR versus BM25 ($p{=}.46$); the two pairings against
dense cross serving chains and so inherit the drift caveat above, which is why
the same-chain MMR--BM25 null is the cleanest of the three. The payload-aware
key moves no writer significantly under the pooled diagnostic ($p \ge .35$),
preserving the absence of a richer-writer advantage rather than proving a tie.

\paragraph{Measured token costs.}
Table~\ref{tab:x-tokens} converts the call-level accounting of
Section~\ref{sec:q4} into measured tokens from the raw serving logs. The
default memory block costs $+604$ prompt tokens per question ($+49\%$) for the
$+4.3$pp transfer lift; widening to $k{=}10$ costs $+1{,}092$ ($+89\%$) for
$+5.6$pp. Completion length is unchanged everywhere: memory is purely a
prompt-side recurring cost. Construction costs another 5.96M tokens in the full
account (4.56M marginal repair-side), which yields the 7.5--9.9K-query
token-amortization horizon in Table~\ref{tab:amortization}. The ladder rows give
the writers' true rendered sizes: the
full-trace writer injects $7.8\times$ the verbatim block's tokens and transfers
worst, the sharpest cost--benefit statement of ``verbosity is not payload.''

\paragraph{Card-level retrieval.}
The body summarizes the broad-use result. Joining every raw solve prompt back
to its retrieved cards (exact join, zero unmatched slots across
$\sim$$6{,}800$ card slots) gives the full audit: $72$--$73\%$ of cards appear in at
least one fixed question's window, the top decile of cards accounts for only
$27$--$32\%$ of fix-window appearances, and retrieval frequency has Gini
$0.34$--$0.36$. The transfer gain is thus carried by many database-specific
payloads, not a few golden templates (converging with the $0.7$--$0.9\%$
template overlap of Section~\ref{sec:threats}). Window credit is
co-occurrence, not per-card causation; we read it as distribution evidence
only. At $k{=}1$, $\sim$$18\%$ of cards are never retrieved; $k{=}5$ already
touches $99$--$100\%$ of the bank, so the $k{=}10$ gain does not come from
waking unused cards, and the quarter-bank collapse of Section~\ref{sec:q3} is
the loss of useful cards, not dead weight. Nor does it come from new verbatim
content: tracing the foreign SQL elements of the $20$ pooled questions fixed at
$k{=}10$ but not $k{=}5$, none traces to a deep card (rank $6$--$10$), nine
trace to a card already present at $k{=}5$, and eleven are untraced, so the
deep window's influence is indirect (context composition), individual marginal
flips sit inside the drift band. The claim is therefore carried by the aggregate
W0 endpoint contrast ($k{=}10-k{=}1$: $+3.18$pp, two-stage $p{=}.0028$), not by
any individual deep card or intermediate step.

\paragraph{Stable versus churn fixes.}
The body reports the $25$-to-$5$ stable fix/break tally. Here, ``stable'' means
the same verdict in every seed where a question is held out, at least twice;
the tally is a $5{:}1$ ratio versus $1.9{:}1$ on pooled events, and no question
flips between fix and break across seeds. The net gain therefore concentrates
in reproducible fixes; the memory tax is real but seed-idiosyncratic rather
than systematic.

\paragraph{A three-case gallery.}
Three held-out cases, verbatim from the logs, show the mechanism in both
directions. \emph{(1) Value-format transfer (stable fix).} Asked for notes of a
fundraising event on 2019/9/14, the bare model writes \texttt{event\_date =
'2019/9/14' AND type = 'fundraising'}: wrong table, wrong date format, wrong
casing. With memory, the window contains a repaired card for ``received funds
on 9/9/2019'' whose SQL reads \texttt{income.date\_received = '2019-09-09'};
the solve adopts the table routing, the ISO date, and the capitalised
\texttt{'Fundraising'}, and is correct. \emph{(2) Dialect convention (stable
fix).} A timestamp equality needs the trailing \texttt{.0}
(\texttt{'\ldots 20:29:39.0'}) in this database; the bare model omits it, the
card-bearing arm reproduces it. \emph{(3) Style seepage (stable break).} Asked
for one player's heading accuracy, the bare model's correct join gains a
spurious \texttt{LIMIT 1} under memory; the only \texttt{LIMIT} in its window
is a neighbouring top-5 ranking card, and the gold answer returns multiple
rows. The same grounded specificity that carries fixes (1) and (2) is what
leaks in (3): the fix and break classes coincide, as the body's fix/break
ledger shows in aggregate.

\paragraph{Seepage and adoption, quantified.}
The body reports the upper-bound trace over changed answers. The gallery's
mechanism is a rate, not an anecdote. Extracting the
\emph{foreign} SQL elements of every paired flip (fragments such as value literals,
\texttt{LIMIT} clauses, and identifiers present in the memory-arm answer but
absent from the bare answer) and checking them against the question's
retrieved cards: $58\%$ of all fixes and $49\%$ of all breaks contain a foreign
element that appears verbatim in a window card, and conditioned on the answer
having changed at all, the two rates are the same ($77\%$ vs.\ $75\%$).
Adoption and seepage are one channel used in two directions; the memory tax is
not a separate failure mode but the payload mechanism itself landing on a
question that did not need it. (Verbatim-match tracing is an upper-bound
attribution, since identifiers can also arrive from the schema; that is why we
read the fix/break \emph{symmetry}, not the absolute rates.)

\paragraph{Substitution, tested at question level.}
The body reports the repair overlap; here is the full question-level test.
Two body sentences assert that test-time-compute methods overlap on the same
errors; both survive a direct question-level test. Of the questions memory
fixes on the greedy base, the memory-less repair loop repairs $59$--$67\%$
(mean $62\%$) against a $26\%$ base rate on comparable failures, a $2.4\times$
enrichment supporting ``their gains substitute rather than stack''
(Section~\ref{sec:q4}). Independently, chain-of-thought alone already solves
$39$--$48\%$ (mean $43\%$) of memory's greedy-base fixes, against a $19.5\%$
base rate ($2.2\times$), which is exactly the preferential absorption behind
the smaller (but surviving) memory lift on the CoT base
(Section~\ref{sec:q5}). Had these overlaps matched their base rates, the
substitution reading would have been falsified.

\section{Prompts, Card Templates, and a Worked Example}
\label{sec:x-protocol}
\suppressfloats[t]

\begin{algorithm}[t]
\caption{From test-time episodes to stored memory. A verification signal $v$
controls storage (Section~\ref{sec:framework-deploy}); the gold SQL is never
an argument to the writer and never enters a card.}
\label{alg:r2m}
\begin{algorithmic}[1]
\Require solver $S$, writer $W$, source $T$ with grounding signal $g$, width $k$,
verification signal $v$
\State $M_d \gets \emptyset$ for every database $d$
\For{$(q, d)$ in the collection stream}  \Comment{in deployment: the serving loop}
  \State $\hat{y} \gets S(q, d)$
  \If{$v(\hat{y})$ accepts} bank $(q,\hat{y},\hat{y},\varnothing)$ \Comment{origin-contrast arm only; the default bank stores repairs}
  \Else
    \State $e = (q, \hat{y}, y^{+}, \tau) \gets T_g(q, d, \hat{y})$ under budget $K$
    \If{$v(y^{+})$ accepts} $M_d \gets M_d \cup \{W(e)\}$
    \EndIf
  \EndIf
\EndFor
\Statex \emph{Evaluation} (strictly separated settings):
\State transfer: $P_M \gets$ EX of $S\big(q' \,\big|\, \rho_k(q'; M_{d'})\big)$
  over held-out $(q', d')$; with $P_0, P_K$ on that same split, report CR as
  transfer lift / on-demand repair headroom (Eq.~\ref{eq:cr})
\State replay: the same one-shot consumer over banked questions, self-card eligible
\State cross-question retention: repeat after removing each question's self-card
\end{algorithmic}
\end{algorithm}

\paragraph{Reproducibility.}
All models are served locally behind an OpenAI-compatible endpoint on a single
multi-GPU host; solving uses a $2{,}048$-token output budget within the model's long
context window, and retrieval a $1{,}024$-dimensional Qwen3-Embedding-0.6B index held
in memory. Every result derives from complete per-question logs: we record the
raw request and response of each model call, the fully assembled solve prompt, and
each repair trajectory. The open-source repository is available at
\url{https://github.com/ai-jiaqian/text-to-sql-memory-crystallization}. It provides the solver,
repair, memory, retrieval, and evaluation code; a locked environment; tests;
frozen aggregate result ledgers; and scripts that regenerate the three result
figures and the main evidence tables. The full BIRD databases, our
quality-controlled split, and raw model transcripts are not redistributed;
the artifact documents this boundary and provides commands for an independent
replication on public BIRD Mini-Dev.

\input{tables/xtabA1_data}

\paragraph{Data coverage.}
Table~\ref{tab:x-data} lists per-database held-out coverage. In total, the study
settles more than $120$ transfer evaluations and $63$ replay/retention evaluations locally
(one evaluation = one configuration $\times$ one seed over a full split), plus
the collection passes that produced every bank. Only the four designated
contrasts of Section~\ref{sec:setup} are read confirmatorily; all other cells
are diagnostics under that multiplicity discipline.

\paragraph{Solver and repair prompts.}
All prompts below are transcribed verbatim from the supplementary
implementation; the retrieved-experience block is the only part that
varies across memory arms, and it is delimited so that provenance is never
leaked. The default solver is single-shot and greedy; an optional
chain-of-thought variant~\cite{wei2022cot,zhou2023leasttomost}, gated by an
environment flag and applied identically across arms, is not used for any
headline number.

\begin{scriptsize}
\begin{verbatim}
SOLVER -- system prompt (default, single-shot):
  You are an expert data analyst who writes correct
  SQLite SQL. Given a database schema and a question,
  output ONE SQLite SELECT query that answers it.
  Rules:
  - Use ONLY the tables and columns in the schema.
  - Quote identifiers with spaces/special chars.
  - Read-only: a single SELECT (or WITH ... SELECT);
    no INSERT/UPDATE/DELETE/DDL.
  - Return exactly the columns the question asks for.
  - Output the SQL inside a ```sql ... ``` block only.

SOLVER -- user message (assembled by build_prompt):
  [Database schema]
  <linked schema>
  [Relevant experience -- verified facts and rules]
  <retrieved cards, under a writer-specific header>
  [Question]
  <question, incl. the BIRD evidence hint>
  Write the SQLite query now, inside a ```sql block.

PROBE proposal -- system prompt (source T3):
  You are debugging a SQLite query judged INCORRECT.
  You do NOT have the correct answer -- only that your
  query is wrong. Diagnose by interrogating the DB with
  small READ-ONLY probe queries.
  [user adds schema, question, previous SQL, what it
   returned, then: "Propose 1-4 READ-ONLY SQLite probe
   queries that would reveal your mistake ... Output
   EACH probe in its own ```sql block. Do not write the
   final answer yet."]

REVISE -- system prompt (probe-grounded, T3):
  You are correcting a SQLite query. Use what the DB
  probes revealed to fix the specific mistake. You still
  cannot see the correct answer. Output ONE corrected
  query.

REVISE -- system prompt (one-bit, source T2):
  You are correcting a SQLite query. Output ONE
  corrected query.
  [the only feedback is the fixed sentence: "The
   previous SQL did not answer the question correctly.
   Please revise it."]

CAPSULE writer -- system prompt (writer W5):
  You write a compact Text2SQL repair memory. Use only
  the provided wrong and repaired SQL. Output JSON only
  with keys trigger, applicability, negative_guard,
  lesson.
  [user gives question, evidence, initial wrong SQL,
   final repaired SQL, and probe reports; a consistency
   fence then blanks any field naming a table, column,
   or value absent from the episode.]
\end{verbatim}
\end{scriptsize}

\paragraph{Card templates and injection headers.}
Retrieved cards enter under writer-specific headers, kept neutral so that no
writer receives extra instruction: W0
\texttt{[Similar solved questions and their final SQL]}; diff
\texttt{[Repair diffs ...]}; diff+guard
\texttt{[Repair diffs with applicability guards ...]}; capsule
\texttt{[Repair capsules ...]}. The origin contrast of Section~\ref{sec:q2} uses
the same neutral W0 header for repair, success, and union banks. The W4 guard is
a fixed lookup keyed by the AST-derived failure mode, not model-written. For
example, the \texttt{value-literal} mode maps to ``Apply only when filtering on
this column; verify the literal actually exists in the column first,'' and
\texttt{group-by} to ``Add GROUP BY only when the answer is per
entity/category''; the full table of fourteen guards is in the implementation.

\paragraph{A worked example: one episode, one ladder.}
The writers differ only in how they render \emph{the same} episode. Below is
one real collection episode (BIRD \texttt{california\_schools}, seed 42; failure
modes \texttt{value-literal}, \texttt{column-ref}) and its card under each writer.
The grounding is visible: a database probe reveals that the label ``Community
Day'' lives in the \texttt{School}/\texttt{DOCType} columns, not in the grade-span
columns the first attempt filtered on: exactly the local, database-specific
content identified by the controlled study as the main carrier of transfer.

\begin{scriptsize}
\begin{verbatim}
EPISODE
Question:
  How many active and closed District Community Day
  Schools are there in the county of Alpine?

First attempt (WRONG):
  SELECT COUNT(*) FROM schools WHERE County = 'Alpine'
    AND (StatusType='Active' OR StatusType='Closed')
    AND (GSoffered LIKE '%Community Day School%'
         OR GSserved LIKE '%Community Day School%')

Database probe (T3) and what it revealed:
  SELECT DISTINCT StatusType, COUNT(*) FROM schools
    WHERE County='Alpine' GROUP BY StatusType
    -> Active|5  Closed|11  Merged|1
  SELECT School, GSoffered, GSserved FROM schools
    WHERE County='Alpine' LIMIT 20
    -> "Community Day" appears in School / DOCType,
       NOT in GSoffered / GSserved

Repaired (CORRECT):
  SELECT COUNT(*) FROM schools WHERE County = 'Alpine'
    AND (StatusType='Active' OR StatusType='Closed')
    AND (School LIKE '%Community Day%'
         OR DOCType LIKE '%Community Day%')

------ the same episode, as each writer stores it ------

W0 verbatim:
  Q: How many active and closed District Community Day
     Schools are there in the county of Alpine?
  ```sql
  SELECT COUNT(*) FROM schools WHERE County='Alpine'
   AND (StatusType='Active' OR StatusType='Closed')
   AND (School LIKE '%Community Day%'
        OR DOCType LIKE '%Community Day%')
  ```

W3 diff+mode:
  Past repair (failure mode: value-literal, column-ref):
  Q: How many active and closed District Community Day ...
  - Wrong: ...(GSoffered LIKE '%Community Day School%'
           OR GSserved LIKE '%Community Day School%')
  - Fixed: ...(School LIKE '%Community Day%'
           OR DOCType LIKE '%Community Day%')
  - Clause: WHERE

W3+A anchor:  = W3 diff+mode, then:
  - Final SQL:
  ```sql
  SELECT COUNT(*) FROM schools WHERE County='Alpine'
   AND ... School LIKE '%Community Day%' OR DOCType
   LIKE '%Community Day%' ...
  ```

W4 guard:  = W3 diff+mode, then (fixed template, no LLM):
  - Guard: Apply only when filtering on this column;
    verify the literal actually exists in the column
    first.

W5 capsule:  = W3 diff+mode, then (LLM-distilled, fenced):
  - Trigger: Question asks for 'Community Day Schools'
    but initial SQL filters on 'GSoffered'/'GSserved'.
  - Apply-when: the target school type is identified by
    name patterns in 'School'/'DOCType', not grade-span
    columns.
  - Do-NOT-apply-when: the type is defined by grade
    ranges in 'GSoffered'/'GSserved', or the schema
    lacks those columns.
  - Lesson: school-type classifications are often stored
    in the school name or document-type fields, not
    grade-span columns.
\end{verbatim}
\end{scriptsize}

\noindent This is the object behind the writer rows of
Figure~\ref{fig:decisions}: W0 has the best
transfer point estimate and the richer renderings show no reliable improvement,
while the capsule's explicit apply/do-not-apply conditions let it replay the
once-failed question every time when its own card is eligible. For
this episode the consistency fence kept all four capsule fields; on cards whose
free-text strays from the episode it blanks the offending field, yet capsule
exact-query replay stays at $100\%$; without the self-card, capsule and verbatim
converge near $56\%$ (Table~\ref{tab:cr-main}).

%% file: tables/table3_amortization.tex
\begin{table}[t]
\caption{Call-level and lifecycle cost (Qwen3.5-27B, three seeds, held-out
$n{=}461$). Memory moves correctness earlier but does not raise the repaired
endpoint. The 59\%/80\% result is a call-level view; the lower panel accounts for
construction and recurring prompt tokens. The reported horizon is token-denominated,
not a full economic break-even. Full construction also executes about 4.3K
database probes per seed; embedding, latency, and dollar cost remain separate.}
\label{tab:amortization}
\centering\footnotesize
\begin{tabular}{@{}lccc@{}}
\toprule
Configuration & First attempt (\%) & Final (\%) & LLM calls/q \\
\midrule
no memory $+$ 1 repair round & 61.46 & 70.43 & 1.77 \\
\textbf{W0 memory $+$ 1 repair round} & \textbf{64.14} & 69.99 & \textbf{1.72} \\
no memory $+$ 3 repair rounds ($P_K$) & 62.19 & 71.95 & 2.92 \\
\midrule
\multicolumn{4}{@{}l}{\emph{Lifecycle token account}} \\
Full / marginal construction & \multicolumn{3}{c}{5.96M / 4.56M tokens per seed} \\
Recurring prompt overhead ($k{=}5$) & \multicolumn{3}{c}{$+604$ tokens/query ($+49\%$)} \\
Tokens per pp per query & \multicolumn{3}{c}{memory 139 / serve-repair 359 ($2.6\times$)} \\
Amortization horizon & \multicolumn{3}{c}{7.5K marginal / 9.9K full future queries} \\
\bottomrule
\end{tabular}
\end{table}

%% file: figures/fig4_consumption.tex
\begin{figure*}[t]
\centering
\includegraphics[width=\textwidth]{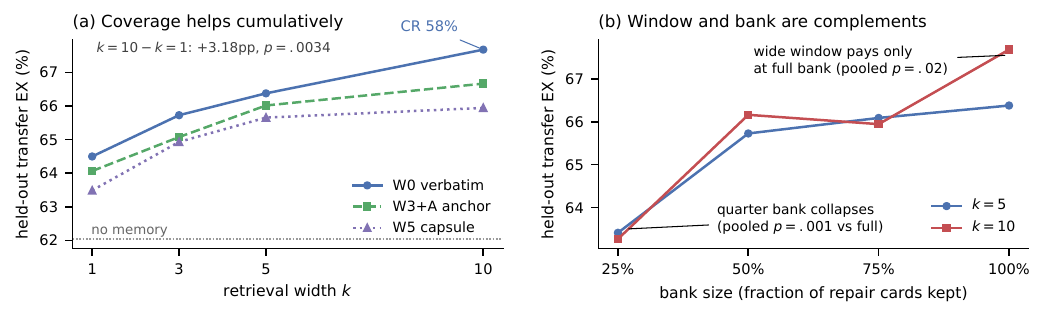}
\caption{\textbf{Retrieval coverage and bank size are complementary.}
\textbf{(a)}~For W0, increasing width from $k{=}1$ to $k{=}10$ yields a cumulative
$+3.18$pp under the two-stage test (CI $[+1.09,+5.35]$, $p{=}.0028$), and W0
reaches a same-source CR of $58\%$. Individual increments are not significant
hierarchically, so the figure supports an endpoint cumulative claim rather than
stepwise monotonicity. Richer writers do not catch W0 at the wide endpoint.
\textbf{(b)}~At the default width $k{=}5$, half the bank has nearly the same
the full bank while a quarter collapses; widening to $k{=}10$ pays most clearly
at the full bank. The panel's $p$ values are pooled diagnostics; the main
inferential claim is the hierarchical endpoint contrast in panel (a).}
\Description{Panel a compares transfer across retrieval widths and memory
writers. Panel b compares bank fractions at retrieval widths five and ten.}
\label{fig:consumption}
\end{figure*}

%% file: tables/table4_crosssections.tex
\begin{table}[t]
\caption{Cross-sections with different evidential strength. The CR result is
replicated over three seeds at two Qwen scales. The four additional family rows
are single-seed preliminary probes without held-out repair-headroom denominators:
all show positive W0 transfer and high exact replay. On Spider (two seeds),
replay ordering repeats while transfer deltas are small, consistent with
Spider's $79$--$80\%$ no-memory floor: the repair headroom measured by CR
is largely absent. Gemma4-E4B's anchored-writer preference is also a
single-seed observation. Full details appear in the Appendix.}
\label{tab:crosssections}
\centering\footnotesize
\setlength{\tabcolsep}{3.4pt}
\resizebox{\columnwidth}{!}{%
\begin{tabular}{@{}lccccc@{}}
\toprule
Model (family) & first-try & repair rate & exact replay $\varnothing\!\to\!$W0 (\%) & $\Delta$W0 transfer & CR \\
\midrule
Qwen3.5-27B & .620 & 28.5\% & 2 $\to$ 96 & $+4.3$pp & 44.4\% \\
Qwen3.5-9B & .499 & 27.8\% & 2 $\to$ 96 & $+6.2$pp & 49.5\% \\
Gemma4-E4B & .519 & 27.1\% & 0 $\to$ 89 & $+6.5$pp & --- \\
Gemma4-31B & .616 & 35.9\% & 1 $\to$ 93 & $+5.2$pp & --- \\
gpt-oss-20b & .527 & 25.6\% & 8.5 $\to$ 95 & $+3.3$pp & --- \\
Hunyuan-A13B & .399 & 13.8\% & 3 $\to$ 97 & $+6.9$pp & --- \\
\midrule
\multicolumn{6}{@{}l}{\emph{Spider (Qwen3.5-27B)}: transfer none/W0/anchor $\cdot$ retention none$\to$W0$\to$anchor} \\
seed 42 & \multicolumn{2}{c}{78.8 / 80.1 / 80.4} & \multicolumn{3}{c}{.03 $\to$ .78 $\to$ .97} \\
seed 7 & \multicolumn{2}{c}{79.7 / 80.1 / 80.7} & \multicolumn{3}{c}{.00 $\to$ .75 $\to$ 1.00} \\
\bottomrule
\end{tabular}
}
\end{table}

%% file: tables/xtabB0_pk.tex
\begin{table}[t]
\caption{Held-out on-demand repair headroom used to normalize CR:
measurement-only oracle-repair pass whose trajectories are never banked. The 9B gap spans $2.8$pp across
seeds against the 27B gap's $0.4$pp.}
\label{tab:x-pk}\centering\footnotesize
\begin{tabular}{@{}lcccc@{}}
\toprule
Model / seed & floor (pk stack) & $P_K$ & gap (pp) & $n$ \\
\midrule
27B / 42 & 61.6 & 71.6 & 9.98 & 461 \\
27B / 7 & 63.3 & 72.9 & 9.54 & 461 \\
27B / 13 & 61.6 & 71.4 & 9.76 & 461 \\
9B / 42 & 51.6 & 63.3 & 11.71 & 461 \\
9B / 7 & 53.6 & 68.1 & 14.53 & 461 \\
9B / 13 & 50.1 & 62.9 & 12.80 & 461 \\
\bottomrule
\end{tabular}
\end{table}

%% file: tables/xtabB1_writers.tex
\begin{table}[t]
\caption{Complete W0--W5 writer ladder (Qwen3.5-27B, seeds 42/7/13).
Both columns are writer-level three-seed means from the same frozen-bank
evaluation chain. Exact-query replay permits the originating card; cross-question
retention after removing it is reported separately.}
\label{tab:x-writers}\centering\footnotesize\setlength{\tabcolsep}{3pt}
\begin{tabular}{@{}lcc@{}}
\toprule
Writer & held-out transfer EX (\%) & exact-query replay (\%) \\
\midrule
W0 verbatim & 67.2 & 96.1 \\
W1 full trace & 65.2 & 99.7 \\
W2 bare diff & 65.9 & 93.5 \\
W3 diff+mode & 65.6 & 91.6 \\
W4 diff+guard & 65.8 & 89.8 \\
W5 capsule & 65.9 & 99.7 \\
\bottomrule
\end{tabular}
\end{table}

%% file: tables/xtabB2_sources.tex
\begin{table}[t]
\caption{Per-seed expansion of the source panel (equal-card matched
contrasts; card counts per seed; transfer EX, held-out transfer/headroom CR, exact-query replay, \%).
Subsample dirs contain repaired lines only, hence counts equal their
source's yield. All T4 rows use verified storage; the ungated control is
Table~\ref{tab:x-matched-admission}.}
\label{tab:x-sources}\centering\footnotesize\setlength{\tabcolsep}{2.6pt}
\begin{tabular}{@{}lcccc@{}}
\toprule
Source & cards (42/7/13) & EX (42/7/13) & CR (42/7/13) & Replay \\
\midrule
T1 resample & 59/63/61 & 63.3/63.8/64.0 & 19.6/6.8/24.4 & 98.3/100.0/100.0 \\
T3sub (T1-matched) & 59/63/60 & 66.6/67.0/66.8 & 52.2/40.9/53.3 & 93.2/98.4/95.0 \\
T2 one bit & 51/58/55 & 65.1/64.0/63.3 & 37.0/9.1/17.8 & 98.0/100.0/98.2 \\
T3sub (T2-matched) & 51/58/55 & 65.9/65.9/66.2 & 45.7/29.5/46.7 & 96.1/98.3/92.7 \\
T4 self-vote & 46/50/49 & 64.9/64.9/64.2 & 34.8/18.2/26.7 & 100.0/96.0/100.0 \\
T3sub (T4-matched) & 46/50/49 & 65.1/66.2/63.1 & 37.0/31.8/15.6 & 97.8/98.0/89.8 \\
\bottomrule
\end{tabular}
\end{table}

%% file: tables/xtabB3_9b.tex
\begin{table}[t]
\caption{Per-seed 9B writer panel (exact-query replay is self-card eligible). The wide CR spread tracks the 9B
denominator variance (Table~\ref{tab:x-pk}); we therefore claim only
that transfer does not degrade at 9B, not a scale trend.}
\label{tab:x-9b}\centering\footnotesize\setlength{\tabcolsep}{3pt}
\begin{tabular}{@{}lccc@{}}
\toprule
Writer & EX (42/7/13) & CR (42/7/13) & Replay (42/7/13) \\
\midrule
none & 51.2/53.8/50.5 & ---/---/--- & 2.7/1.4/2.0 \\
W0 verbatim & 59.7/57.3/57.3 & 72.2/23.9/52.5 & 96.7/96.5/93.9 \\
W3+A anchor & 56.0/58.4/56.6 & 40.7/31.3/47.5 & 96.7/97.9/95.9 \\
W5 capsule & 55.5/56.4/58.1 & 37.0/17.9/59.3 & 98.0/97.9/98.6 \\
\bottomrule
\end{tabular}
\end{table}

%% file: tables/xtabC1_kwriter.tex
\begin{table}[t]
\caption{Full writer $\times$ retrieval-width grid (held-out transfer
EX \%, per seed). Means are ordered with width, but seed-level steps are not.
The primary two-stage result is W0 $k{=}10-k{=}1$: $+3.18$pp,
$p{=}.0028$; individual increments are not significant hierarchically.}
\label{tab:x-kwriter}\centering\footnotesize
\begin{tabular}{@{}lccccc@{}}
\toprule
Writer & $k$ & s42 & s7 & s13 & mean \\
\midrule
W0 verbatim & 1 & 64.4 & 63.6 & 65.5 & 64.5 \\
 & 3 & 62.5 & 67.2 & 67.5 & 65.7 \\
 & 5 & 65.9 & 66.8 & 66.4 & 66.4 \\
 & 10 & 67.9 & 67.5 & 67.7 & 67.7 \\
\addlinespace
W3+A anchor & 1 & 63.8 & 63.6 & 64.9 & 64.1 \\
 & 3 & 62.5 & 67.2 & 65.5 & 65.1 \\
 & 5 & 65.1 & 66.4 & 66.6 & 66.0 \\
 & 10 & 64.9 & 67.9 & 67.2 & 66.7 \\
\addlinespace
W5 capsule & 1 & 62.5 & 63.6 & 64.4 & 63.5 \\
 & 3 & 61.8 & 65.9 & 67.0 & 64.9 \\
 & 5 & 64.0 & 65.9 & 67.0 & 65.7 \\
 & 10 & 63.6 & 66.8 & 67.5 & 65.9 \\
\addlinespace
\bottomrule
\end{tabular}
\end{table}

%% file: tables/xtabC2_banksize.tex
\begin{table}[t]
\caption{Full bank-size $\times$ window grid (W0, held-out transfer EX
\%, per seed; card counts per seed in the first column). Half vs.\ full
bank at $k{=}5$: pooled $p{=}.41$; quarter vs.\ full: pooled $p{=}.001$;
$k{=}10$ vs.\ $k{=}5$ at full bank: pooled $p{=}.020$. These are supporting
diagnostics; the hierarchical width claim is Table~\ref{tab:x-kwriter}.}
\label{tab:x-banksize}\centering\footnotesize
\begin{tabular}{@{}lcccccc@{}}
\toprule
Bank & cards (42/7/13) & $k$ & s42 & s7 & s13 & mean \\
\midrule
25\% & 27/28/28 & 5 & 62.3 & 64.2 & 63.8 & 63.4 \\
 &  & 10 & 62.5 & 63.8 & 63.6 & 63.3 \\
50\% & 60/62/62 & 5 & 65.7 & 65.5 & 65.9 & 65.7 \\
 &  & 10 & 65.5 & 66.6 & 66.4 & 66.2 \\
75\% & 90/92/92 & 5 & 66.2 & 66.6 & 65.5 & 66.1 \\
 &  & 10 & 65.3 & 67.2 & 65.3 & 65.9 \\
100\% & 124/129/128 & 5 & 65.9 & 66.8 & 66.4 & 66.4 \\
 &  & 10 & 67.9 & 67.5 & 67.7 & 67.7 \\
\bottomrule
\end{tabular}
\end{table}

%% file: tables/xtabB4_controls.tex
\begin{table}[t]
\caption{Retrieval controls, per seed (W0 bank, $k{=}5$, held-out transfer EX
\%). Both cross-database arms are below the no-memory floor in every seed
(all-foreign vs.\ floor $p{=}.006$; schema-similar vs.\ all-foreign $p{=}.30$,
i.e.\ similarity does not rescue); the random-retrieval arm sits between floor
and top-$k$ (vs.\ floor $p{=}.020$, vs.\ top-$k$ $p{=}.010$).}
\label{tab:x-controls}\centering\footnotesize
\begin{tabular}{@{}lccccc@{}}
\toprule
Arm & s42 & s7 & s13 & mean & CR \\
\midrule
no memory (floor) & 61.4 & 63.1 & 61.6 & 62.0 & --- \\
cross-DB pool (all-foreign) & 59.7 & 59.9 & 59.4 & 59.7 & $-24.6\%$ \\
cross-DB pool (schema-similar) & 59.7 & 62.0 & 59.9 & 60.5 & $-15.5\%$ \\
random same-DB & 65.1 & 64.9 & 63.1 & 64.4 & $23.6\%$ \\
top-$k$ same-DB (default) & 65.9 & 66.8 & 66.4 & 66.4 & $44.4\%$ \\
\bottomrule
\end{tabular}
\end{table}

%% file: tables/xtabF2_permuted_content.tex
\begin{table*}[t]
\caption{Correspondence control ladder on held-out BIRD transfer (EX \%, W0,
$k{=}5$). Aligned and permuted are paired within one rerun; the no-memory and
random rows reuse controls from the same split. Permutation preserves the local
questions, SQL multiset, card count, retrieval, and prompt form while
mispairing every SQL. The permuted bank improves accuracy by $+3.91$pp (CI
approximately $[+1.4,+6.9]$, $p{<}.01$), retaining $73\%$ of the aligned
bank's $+5.36$pp lift. Restoring correct pairing adds $+1.45$pp, but its
interval includes zero (CI approximately $[-0.5,+3.2]$, $p{=}.14$). The fresh
aligned rerun differs from its earlier run of the same configuration by only
$+0.22$pp. These rows are separate controls, not a complete factorial design.}
\label{tab:x-permuted}
\centering\footnotesize
\begin{tabular}{@{}lccrrrrr@{}}
\toprule
Arm & Similarity targeted & Q--SQL aligned & s42 & s7 & s13 & Mean & $\Delta$ none \\
\midrule
No memory & --- & --- & 61.39 & 62.83 & 61.39 & 61.87 & --- \\
Random same-DB & no & yes & 65.08 & 65.00 & 63.12 & 64.40 & $+2.53$ \\
Permuted same-DB & yes & no & 64.86 & 67.17 & 65.29 & 65.78 & $+3.91$ \\
Aligned same-DB & yes & yes & 66.38 & 68.04 & 67.25 & 67.22 & $+5.35$ \\
\bottomrule
\end{tabular}
\end{table*}

%% file: tables/xtabC5_srcaxis.tex
\begin{table}[t]
\caption{Origin contrast, per seed (held-out EX \%; deltas vs.\ the
run's own no-memory floor). Per-item difference sets: repair-only
correct 31/28/24 vs.\ success-only correct 25/26/30 per seed; only one
question is stably unique to either side across all seeds; pooled
McNemar $p{\approx}.94$.}
\label{tab:x-srcaxis}\centering\footnotesize
\begin{tabular}{@{}lcccc@{}}
\toprule
Bank & s42 & s7 & s13 & mean $\Delta$ (pp) \\
\midrule
none & 61.4 & 62.7 & 61.4 & --- \\
first-try-correct & 64.6 & 67.2 & 68.5 & +4.99 \\
repaired & 65.9 & 67.7 & 67.2 & +5.13 \\
union & 66.8 & 67.2 & 68.3 & +5.64 \\
\bottomrule
\end{tabular}
\end{table}

%% file: tables/xtabF1_matched_admission.tex
\begin{table*}[t]
\caption{Per-seed evidence for the matched-count episode-source control and the
verification boundary (BIRD, Qwen3.5-27B, $k{=}5$). Success banks are downsampled
per database to the repair-bank count in two deterministic draws. Roughly 127
ordinary verified successes retain most transfer; repair has a higher point estimate
in all six matched cells, but the two-stage repair advantage is not significant
($+1.30/+2.32$pp, $p{=}.304/.073$). In the verification panel, the otherwise matched
verified--ungated T4 contrast is $+4.85$pp (two-stage $p{=}.0064$).}
\label{tab:x-matched-admission}
\centering\footnotesize
\begin{tabular}{@{}lccccc@{}}
\toprule
Configuration & seed 42 & seed 7 & seed 13 & mean & lift over none \\
\midrule
\multicolumn{6}{@{}l}{\emph{Matched-count episode-source control (transfer EX, \%)}} \\
none & 61.39 & 62.83 & 61.39 & 61.87 & --- \\
success full ($\sim$635 cards) & 64.64 & 67.39 & 68.55 & 66.86 & $+4.99$pp \\
success matched, draw 0 ($\sim$127) & 65.29 & 65.87 & 65.94 & 65.70 & $+3.83$pp \\
success matched, draw 1 ($\sim$127) & 63.99 & 65.43 & 64.64 & 64.69 & $+2.82$pp \\
repair matched ($\sim$127) & 65.94 & 67.83 & 67.25 & 67.00 & $+5.14$pp \\
\midrule
\multicolumn{6}{@{}l}{\emph{Self-vote verification control (transfer EX, \%)}} \\
self-vote, unverified storage & 60.52 & 59.35 & 59.65 & 59.84 & $-2.03$pp \\
self-vote, verified storage & 64.86 & 65.00 & 64.21 & 64.69 & $+2.82$pp \\
\bottomrule
\end{tabular}
\end{table*}

%% file: tables/xtabC3_families.tex
\begin{table*}[t]
\caption{Family cross-section, full detail (transfer EX and exact-query replay,
\%; Qwen rows are three-seed means, other rows single-seed minimal
probes by design). All preliminary estimates are positive, but only the two Qwen
rows support a multi-seed scale replication; magnitudes and the best point
estimate vary.}
\label{tab:x-families}\centering\footnotesize
\begin{tabular}{@{}lcccccccc@{}}
\toprule
Model & first-try & repair rate & none & W0 & W3+A & floor replay & W0 replay & seeds \\
\midrule
Qwen3.5-27B (Alibaba) & .620 & 28.5\% & 62.0 & 66.4 & 66.0 & 1.8 & 96.3 & 3 \\
Qwen3.5-9B (Alibaba) & .499 & 27.8\% & 51.8 & 58.1 & 57.0 & 2.0 & 95.7 & 3 \\
Gemma4-E4B (Google) & .519 & 27.1\% & 51.0 & 57.5 & 60.3 & 0.0 & 89.3 & 1 \\
Gemma4-31B (Google) & .616 & 35.9\% & 61.8 & 67.0 & 65.9 & 0.7 & 93.2 & 1 \\
gpt-oss-20b (OpenAI) & .527 & 25.6\% & 54.9 & 58.1 & 57.0 & 8.5 & 94.6 & 1 \\
Hunyuan-A13B (Tencent) & .399 & 13.8\% & 39.5 & 46.4 & --- & 3.3 & 96.7 & 1 \\
\bottomrule
\end{tabular}
\end{table*}

%% file: tables/xtabC7_perdb.tex
\begin{table}[t]
\caption{Per-database heterogeneity of W0 transfer (three seeds pooled).
Ten of eleven databases gain; the top three carry 48\% of the pooled
net gain; a database's card count barely predicts its lift (Pearson
$r{=}.18$). Databases are anonymized to their BIRD names.}
\label{tab:x-perdb}\centering\footnotesize\setlength{\tabcolsep}{3.5pt}
\begin{tabular}{@{}lccccc@{}}
\toprule
Database & $n$ (3 seeds) & cards & floor \% & $\Delta$ (pp) & net q \\
\midrule
toxicology & 132 & 41 & 58.3 & +12.12 & +16 \\
superhero & 117 & 28 & 84.6 & +6.84 & +8 \\
california\_schools & 81 & 37 & 24.7 & +6.17 & +5 \\
formula\_1 & 156 & 37 & 57.1 & +4.49 & +7 \\
european\_football\_2 & 117 & 21 & 66.7 & +4.27 & +5 \\
financial & 96 & 15 & 19.8 & +4.17 & +4 \\
thrombosis\_prediction & 147 & 35 & 59.2 & +3.40 & +5 \\
codebase\_community & 168 & 60 & 75.0 & +2.98 & +5 \\
student\_club & 141 & 45 & 77.3 & +2.84 & +4 \\
card\_games & 171 & 39 & 65.5 & +2.34 & +4 \\
debit\_card\_specializing & 57 & 23 & 73.7 & -5.26 & -3 \\
\bottomrule
\end{tabular}
\end{table}

%% file: tables/table1_instantiation.tex
\begin{table}[t]
\caption{Controlled factors and interventions. The solver, model, split, prompt
injection point, and evaluation remain fixed; each intervention changes one
memory decision against the stated default. Main contrasts use three paired
seeds, while preliminary cross-section probes are labelled separately.}
\label{tab:instantiation}
\centering\footnotesize
\resizebox{\columnwidth}{!}{%
\begin{tabular}{@{}p{0.2\linewidth}p{0.24\linewidth}p{0.56\linewidth}@{}}
\toprule
Factor & Default & Controlled interventions \\
\midrule
Episode source & T3 repair &
T1 resampling; T2 correctness bit; T4 self-vote; first-try successes;
matched-count banks \\
\addlinespace
Verification & Oracle verified &
Verified vs. ungated self-vote; synthetic verification noise \\
\addlinespace
Card format & W0 verbatim &
Trace; SQL diff; failure mode; anchor; guard; natural-language capsule \\
\addlinespace
Retrieval & Same-DB top-$5$ &
$k\in\{1,3,5,10\}$; bank size; random, dense, lexical, payload-aware,
and diversity retrieval; foreign and schema-similar pools \\
\addlinespace
Correspondence & Aligned Q--SQL &
Within-database SQL derangement preserving questions, SQL multiset, retrieval,
card count, and prompt format \\
\addlinespace
Measurement & Held-out transfer + CR &
Collection-side exact-query replay; cross-question retention; token/call cost;
model scale, family, solver, and benchmark probes \\
\bottomrule
\end{tabular}
}
\end{table}

%% file: tables/table_rq2_mechanism.tex
\begin{table*}[t]
\centering
\caption{\textbf{RQ2 controlled comparisons.} Each row changes one property of
the retrieved memory. Effects are paired changes in held-out transfer with
two-stage 95\% intervals over databases and questions.}
\label{tab:rq2-mechanism}
\small
\setlength{\tabcolsep}{3.5pt}
\renewcommand{\arraystretch}{1.12}
\begin{tabularx}{\textwidth}{@{}>{\raggedright\arraybackslash}p{2.45cm}>{\raggedright\arraybackslash}p{3.15cm}>{\raggedright\arraybackslash}p{3.25cm}>{\raggedright\arraybackslash}p{1.45cm}>{\raggedright\arraybackslash}X@{}}
\toprule
Factor & Controlled contrast & $\Delta$ transfer [95\% CI] & Status & Interpretation \\
\midrule
Database locality & Local aligned - foreign aligned & $+6.73$pp $[+2.69,+12.06]$; $p<.001$ & Supported & Valid foreign examples cannot explain the local gain. \\
Grounded content & Full local card - lesson-only & $+2.68$pp $[+0.17,+5.21]$; $p=.037$ & Supported & SQL and database anchors carry reusable information. \\
Content targeting & Similarity-targeted - random local & $+2.03$pp $[+0.22,+3.84]$; $p=.026$ & Supported & Question similarity acts as a content-relevance filter. \\
Mispaired local content & Mispaired local - no memory & $+3.91$pp $[+1.40,+6.87]$; $p=.001$ & Supported & Exact question-SQL pairing is not required for most transfer. \\
Residual correspondence & Aligned - permuted pairing & $+1.45$pp $[-0.45,+3.23]$; $p=.142$ & CI includes zero & Correct pairing may add a smaller term, but is not required for most transfer. \\
Foreign-card diagnostic & Foreign aligned - no memory & $-2.39$pp $[-5.79,+0.56]$; $p=.133$ & CI includes zero & Foreign cards trend downward, but this comparison alone is not conclusive. \\
\bottomrule
\end{tabularx}
\end{table*}

%% file: tables/table_rq3_decisions.tex
\begin{table*}[t]
\centering
\caption{\textbf{RQ3 memory choices.} The table separates statistically
supported comparisons from estimates whose confidence intervals include zero.
Exploratory format and retrieval sweeps are labeled accordingly.}
\label{tab:rq3-decisions}
\small
\setlength{\tabcolsep}{3.5pt}
\renewcommand{\arraystretch}{1.12}
\begin{tabularx}{\textwidth}{@{}>{\raggedright\arraybackslash}p{1.65cm}>{\raggedright\arraybackslash}p{3.25cm}>{\raggedright\arraybackslash}p{4.05cm}>{\raggedright\arraybackslash}p{1.75cm}>{\raggedright\arraybackslash}X@{}}
\toprule
Choice & Controlled comparison & Result & Status & Interpretation \\
\midrule
Verification & Verified vs. ungated self-vote & $+4.85$pp $[+1.30,+8.06]$; $p=.006$ & Supported & Reliable verification is necessary. \\
Episode source & Repair vs. matched successes (two draws) & $+1.30$pp $[-1.07,+3.78]$; $+2.32$pp $[-0.22,+4.94]$ & CIs include zero & Repair targets failures efficiently, but shows no unique transfer advantage. \\
Card format & Richer W1--W5 vs. verbatim W0 & Point estimates $-1.95$ to $-1.30$pp; all CIs cross zero & No supported gain & W0 has the highest point estimate; format equivalence is not established. \\
Retrieval & Retrieval width $k=10$ vs. $k=1$ & $+3.18$pp $[+1.09,+5.35]$; $p=.003$ & Supported & The $k{=}1$ to $k{=}10$ gain is supported; individual steps are not. \\
Retriever method & Dense, BM25, payload-aware, and MMR after scoping/targeting & Targeted methods span $<0.7$pp & Exploratory & More complex methods show no supported gain beyond database scope and targeting. \\
\bottomrule
\end{tabularx}
\end{table*}

%% file: tables/table_mechmap.tex
\begin{table}[t]
\caption{Mechanisms from prior self-evolving text-to-SQL systems, re-instantiated
as controlled interventions (same pipeline, model, and retrieval;
three seeds). The rows contrast \emph{mechanisms}, not systems: whole-system
scores would entangle every pipeline difference with the memory design.}
\label{tab:mechmap}
\centering\footnotesize
\begin{tabular}{@{}p{0.30\linewidth}p{0.28\linewidth}p{0.30\linewidth}@{}}
\toprule
Mechanism in prior line & Controlled intervention & Measured reading \\
\midrule
Corrected-query memory \cite{memosql2026,lpesql2024} & W0 verbatim & CR $44.4\%$; $+4.3$pp \\
Distilled guidelines \cite{magic2025} & W5 capsule & CR $37.1\%$; exact-query replay $100\%$ \\
\quad guideline w/o grounded anchors & W5 lesson-only & CR $17.3\%$ (same-db) \\
Self-consistency filtering \cite{orange2025} & T4 self-vote source & verified bank $+2.82$pp; ungated bank $-2.03$pp \\
Demonstration retrieval \cite{dailsql2024} & first-try-correct bank & matched-count bank yields $+2.82$--$3.83$pp \\
Naive (untargeted) recall & random same-db draw & CR $23.6\%$, half the targeted effect \\
\bottomrule
\end{tabular}
\end{table}

%% file: tables/xtabD1_audit.tex
\begin{table*}[t]
\caption{Claim audit: comparisons supporting the main claims and their strongest
evidence. Two-stage tests resample databases then unique questions, keeping
seed replicates together; pooled discordant counts are supplementary.}
\label{tab:x-audit}
\centering\footnotesize
\begin{tabular}{@{}lll@{}}
\toprule
Claim (section) & Comparison & Evidence \\
\midrule
Transfer is real (\S\ref{sec:q1}) & W0 vs.\ none & $+4.34$pp; two-stage CI $[1.50, 7.49]$, $p{=}.0034$ \\
Fix/break ledger (\S\ref{sec:q1}) & W0 vs.\ none & fixed 127 / broke 67 (net $+60$ of 1{,}383) \\
Lucky repairs collapse (\S\ref{sec:q2}) & T1 vs.\ matched T3 & per-seed $p{=}.04/.04/.09$ \\
One bit lifts (\S\ref{sec:q2}) & T2 vs.\ matched T3 & CR $21.3$ vs.\ $40.6\%$; $\Delta$EX $-1.9$pp \\
Reliable verification (\S\ref{sec:q3}) & gated T4 vs.\ ungated vote-admit & $+4.85$pp; two-stage CI $[1.30,8.06]$, $p{=}.0064$ \\
Matched successes transfer (\S\ref{sec:q2}) & success@matched d0 vs.\ none & $+3.84$pp; two-stage $p{=}.0018$ \\
Repair shows no unique transfer edge (\S\ref{sec:q3}) & repair vs.\ matched success & $+1.30/+2.32$pp; two-stage $p{=}.304/.073$ \\
No reliable richer-writer gain (\S\ref{sec:q3}) & W1--W5 vs.\ W0 & all lower $1.3$--$2.0$pp; hierarchical CIs cross zero \\
Rich formats trend negative at 9B (\S\ref{sec:q3}) & W3+A / W5 vs.\ W0 & $-1.09$ CI $[-2.53, +0.36]$; $-1.37$ CI $[-2.97, +0.22]$ \\
Width helps cumulatively (\S\ref{sec:q3}) & $k{=}10$ vs.\ $k{=}1$ (W0) & $+3.18$pp; two-stage CI $[1.09,5.35]$, $p{=}.0028$ \\
Half bank $\approx$ full (\S\ref{sec:q3}) & 50\% vs.\ 100\%, $k{=}5$ & $b/c = 42/51$, $p{=}.41$ \\
Quarter bank collapses (\S\ref{sec:q3}) & 25\% vs.\ 100\%, $k{=}5$ & $b/c = 55/96$, $p{=}.001$ \\
Wide window needs full bank (\S\ref{sec:q3}) & $k{=}10$ vs.\ $k{=}5$ @100\% & $b/c = 36/18$, $p{=}.020$ \\
Raw vote precision (\S\ref{sec:framework-deploy}) & vote-selected vs.\ oracle & $145/3{,}718$ attempts $= 3.9\%$ ($3.8$--$4.0$ per seed) \\
Random retrieval keeps half (\S\ref{sec:q3}) & rand vs.\ none / vs.\ W0 & $b/c = 105/73$, $p{=}.020$; $41/69$, $p{=}.010$ \\
Cross-DB pool hurts (\S\ref{sec:q3}) & xdb vs.\ none & $b/c = 51/84$, $p{=}.006$; below floor all seeds \\
Schema similarity no rescue (\S\ref{sec:q3}) & xdbsim vs.\ xdb / vs.\ none & $b/c = 62/50$, $p{=}.30$; $42/63$, $p{=}.05$, below floor all seeds \\
Mispaired local bank still transfers (\S\ref{sec:q3}) & permuted vs.\ none & $+3.91$pp; two-stage CI $[1.4,6.9]$, $p{<}.01$ \\
Correspondence CI includes zero (\S\ref{sec:q3}) & aligned vs.\ permuted & $+1.45$pp; two-stage CI $[-0.5,3.2]$, $p{=}.14$ \\
9B W0 lift (\S\ref{sec:q5}) & W0 vs.\ none (9B) & $+6.2$pp, CI $[+4.0, +8.5]$ \\
Writer optimum can flip (\S\ref{sec:q5}) & W3+A vs.\ W0 (E4B) & $+2.8$pp, $p{=}.035$ (one seed) \\
\bottomrule
\end{tabular}
\end{table*}

%% file: tables/xtabE1_tokens.tex
\begin{table}[t]
\caption{Measured solver-call token costs (mean per question, three seeds,
from raw serving logs). Memory cost is entirely prompt-side; completion length
is flat. The ladder rows share one chain (retrieval-matched), so prompt deltas
are the writers' rendered-card sizes as the solver actually sees them.}
\label{tab:x-tokens}
\centering\footnotesize
\begin{tabular}{@{}lrrr@{}}
\toprule
Configuration & Prompt tok & $\Delta$ vs.\ none & Completion tok \\
\midrule
No memory                & $1{,}232$ & ---      & $77$ \\
Verbatim, $k{=}1$        & $1{,}371$ & $+139$   & $79$ \\
Verbatim, $k{=}5$ (default) & $1{,}836$ & $+604$   & $78$ \\
Verbatim, $k{=}10$       & $2{,}324$ & $+1{,}092$ & $78$ \\
\midrule
Ladder, $k{=}5$: verbatim (W0) & $1{,}832$ & $+600$   & $78$ \\
\quad full trace (W1)    & $5{,}938$ & $+4{,}706$ & $79$ \\
\quad bare diff (W2)     & $1{,}913$ & $+681$   & $81$ \\
\quad diff+mode (W3)     & $1{,}953$ & $+721$   & $79$ \\
\quad diff+guard (W4)    & $2{,}075$ & $+842$   & $78$ \\
\quad capsule (W5)       & $2{,}852$ & $+1{,}620$ & $78$ \\
\bottomrule
\end{tabular}
\end{table}

%% file: tables/xtabE2_difficulty.tex
\begin{table}[t]
\caption{The headline transfer contrast (verbatim memory vs.\ no memory),
re-read two ways. Top: under the stricter subset metric, per seed. Bottom:
stratified by BIRD's difficulty labels, pooled over seeds. All cells positive
and significant.}
\label{tab:x-difficulty}
\centering\footnotesize
\begin{tabular}{@{}lrrrr@{}}
\toprule
\multicolumn{5}{@{}l}{\emph{Subset metric (per seed)}} \\
Seed & none & W0 & lift (pp) & $p$ \\
\midrule
42 & $63.8$ & $68.1$ & $+4.3$ & $.008$ \\
7  & $65.9$ & $69.8$ & $+3.9$ & $.036$ \\
13 & $63.6$ & $68.8$ & $+5.2$ & $.004$ \\
\midrule
\multicolumn{5}{@{}l}{\emph{Difficulty strata (EX, pooled)}} \\
Stratum & $n$ pairs & lift (pp) & fix/break & $p$ \\
\midrule
simple      & $775$ & $+3.0$ & $60/37$ & $.025$ \\
moderate    & $404$ & $+6.4$ & $51/25$ & $.004$ \\
challenging & $203$ & $+5.4$ & $16/5$  & $.027$ \\
\bottomrule
\end{tabular}
\end{table}

%% file: tables/xtabA1_data.tex
\begin{table}[t]
\caption{Held-out coverage per database (summed over the three seeds'
$70/30$ splits; each seed holds out 461 of 1{,}534 BIRD questions).}
\label{tab:x-data}\centering\footnotesize
\begin{tabular}{@{}lc@{}}
\toprule
Database & held-out $n$ (3 seeds) \\
\midrule
card\_games & 171 \\
codebase\_community & 168 \\
formula\_1 & 156 \\
thrombosis\_prediction & 147 \\
student\_club & 141 \\
toxicology & 132 \\
european\_football\_2 & 117 \\
superhero & 117 \\
financial & 96 \\
california\_schools & 81 \\
debit\_card\_specializing & 57 \\
\bottomrule
\end{tabular}
\end{table}